\documentclass[utf8]{FrontiersinHarvard} 

\usepackage{url,hyperref,microtype,subcaption}
\usepackage[onehalfspacing]{setspace}

\usepackage{graphics} 
\usepackage{comment}
\usepackage{cite}
\usepackage{enumerate}
\usepackage{amsmath} 
\usepackage{amssymb}  
\usepackage{amsfonts}

\DeclareMathOperator*{\argmax}{argmax}

\usepackage{type1cm}

\usepackage {colortbl,array,xcolor}

\usepackage{algorithm} 
\usepackage{algpseudocode} 
\algrenewcommand{\algorithmicreturn}{\State \textbf{return}}

\usepackage{color}    
\usepackage{umoline} 
\usepackage{proofread} %
\noproofreadmark 

\def\keyFont{\fontsize{8}{11}\helveticabold }
\def\firstAuthorLast{Akira Taniguchi {et~al.}} 
\def\Authors{Akira Taniguchi\,$^{1,*}$, Shuya Ito\,$^{1}$ and Tadahiro Taniguchi\,$^{1}$}
\def\Address{$^{1}$Emergent Systems Laboratory, Ritsumeikan University, 1-1-1 Noji-Higashi, Kusatsu, Shiga 525-8577, Japan}

\def\corrAuthor{Akira Taniguchi}
\def\corrEmail{a.taniguchi@em.ci.ritsumei.ac.jp}

\begin{document}
\onecolumn
\firstpage{1}

\title[Hierarchical Path-planning from Speech Instructions]{Hierarchical Path-planning from Speech Instructions with Spatial Concept-based Topometric Semantic Mapping} 

\author[\firstAuthorLast ]{\Authors} 
\address{} 
\correspondance{} 

\extraAuth{}

\maketitle

\begin{abstract}

Assisting individuals in their daily activities through autonomous mobile robots, especially for users without specialized knowledge, is a significant concern. Specifically, the capability of robots to navigate to destinations based on human speech instructions is crucial.
%
Although robots can take different paths toward the same objective, the shortest path is not always \addspan{the most suitable}.
A preferred approach is to accommodate waypoint specifications flexibly, for the planning of an improved alternative path, even with detours.
Furthermore, robots require real-time inference capabilities.
Spatial representations include semantic, topological, and metric-level representations, each capturing different aspects of the environment.
This study aimed to realize a hierarchical spatial representation using a topometric semantic map and path planning with speech instructions, including waypoints.
%
This paper presents a hierarchical path-planning method called \textit{Spatial Concept-based Topometric Semantic Mapping for Hierarchical Path Planning} (SpCoTMHP), which integrates place connectivity.
This approach provides a novel integrated probabilistic generative model and fast approximate inference with interaction among the hierarchy levels. 
A formulation based on \textit{control as probabilistic inference} theoretically supports the proposed path planning algorithm.
%
We conducted experiments in home environments using the Toyota Human Support Robot on the SIGVerse simulator and in a lab-office environment with the real robot, Albert.
The user issues speech commands that specify the waypoint and goal, such as ``\textit{Go to the bedroom via the corridor.}''
Navigation experiments using speech instruction with a waypoint demonstrated the performance improvement of the SpCoTMHP compared to the baseline, hierarchical path planning method with heuristic path costs (HPP-I), in terms of the weighted success rate at which the robot reaches the closest target and passes the correct waypoints, by 0.590. 
The computation time was significantly accelerated by 7.14 sec with the SpCoTMHP compared to baseline HPP-I in advanced tasks.
Thus, hierarchical spatial representations provide a mutually understandable form for humans and robots, thus enabling language-based navigation tasks.

\tiny
 \keyFont{ \section{Keywords:} control as probabilistic inference, language navigation, hierarchical path planning, probabilistic generative model, semantic map, topological map} 
\end{abstract}

\section{Introduction}
\label{sec:introduction}

%
Autonomous robots are tasked with linguistic interactions, such as navigation, for seamless integration into human environments. 
Navigation using concepts and vocabulary tailored to specific locations learned from human and environmental interactions is a complex challenge for these robots~\citep{taniguchi2015symbol,taniguchi2019langrobo}. 
The robot is required to construct adaptive spatial structures and place semantics from multimodal observations acquired during movement in the environment~\citep{kostavelis2015semantic,Garg2020}. 
This concept is closely linked to the anchoring problem, which concerns the relationship between symbols and sensor observations~\citep{Coradeschi2003,Galindo2005}.
Understanding the specific place or concept to which a word or phrase refers, i.e., its denotations, is crucial.

The motivation for researching this topic stems from the necessity for autonomous robots to operate effectively in human environments. This requires them to understand human language and navigate complex environments accordingly. 
The significance of this research lies in enabling autonomous robots to interact within human environments both effectively and intuitively, thereby assisting users. 
A primary issue in hierarchical path planning is the increase in computational costs due to the complexity of the model, which poses a risk to real-time responsiveness and efficiency. 
Additionally, a challenge with everyday natural language commands provided by users is the existence of specific place names not generally known and the occurrence of different places sharing the same name within an environment.
Therefore, robots need to possess environment-specific knowledge. 
Enhancements in navigation success rates and computational efficiency, especially for tasks involving linguistic instructions, could significantly broaden the applications of autonomous robots. These applications extend beyond home support to include disaster rescue, medical assistance, and more.

Topometric semantic maps, a combination of metric and topological maps with semantics, are helpful for path planning using generalized place units. Thus, they facilitate human-robot linguistic interactions and assist humans. 
One key challenge is the robot's capacity to efficiently construct and utilize these hierarchical spatial representations for interaction tasks.

\begin{figure}[!tb]
  \begin{center}
    \includegraphics[width=0.60\linewidth, clip]{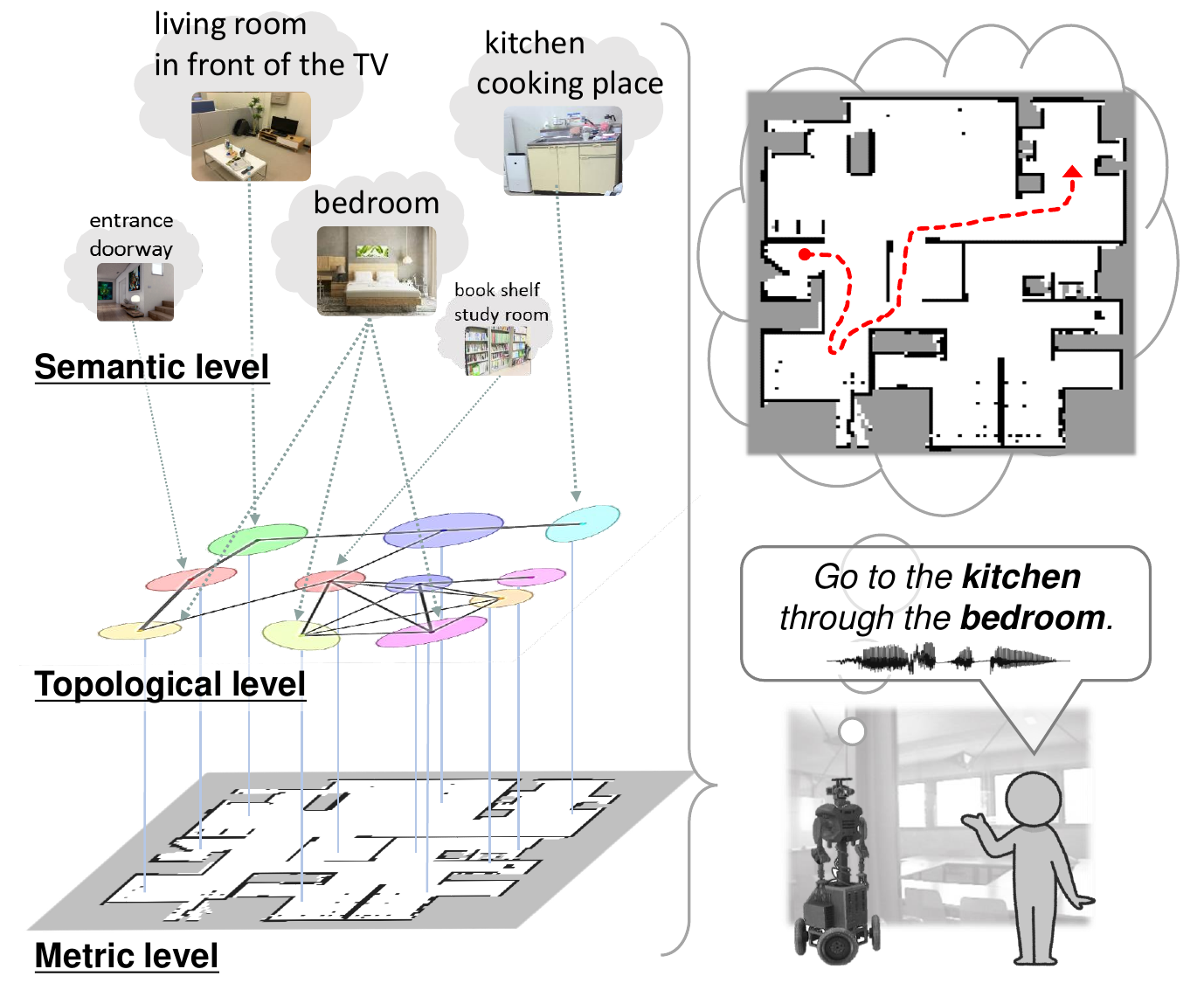}
    \caption{
    Overview.
    Left: Hierarchy of spatial representation with topometric semantic mapping; 
    Right: Path planning from spoken instruction with waypoint and goal specifications.
    }
    \label{fig:overview}
  \end{center}
\end{figure}

Hierarchical spatial representations provide a mutually understandable form for humans and robots to render language-based navigation tasks feasible.
They are generalized appropriately at each level and can accommodate combinations of paths not considered during training.
As shown in Fig.~\ref{fig:overview} (left), this study deals with the three levels of spatial representation: 
(i) \textbf{the semantic level}, which represents place categories associated with various words and abstracted by multimodal observations;
(ii) \textbf{the topological level}, which represents the probabilistic adjacency of places in a graph structure; 
(iii) \textbf{the metric level}, which represents the occupancy grid map and is obtained by simultaneous localization and mapping (SLAM)~\citep{gridbasedfastslam2007}.
In this paper, the term \textit{spatial concepts} refers to semantic-topological knowledge grounded in real-world environments.

%

The main goal of this study was to realize efficient spatial representations and high-speed path planning from human speech instructions specifying waypoints using topological semantic maps that incorporated place connectivity.
This study was conducted in two phases: spatial concept learning and path planning.
\\
\textbf{Spatial concept learning phase}:
In the learning phase, a user guides a robot in the environment, providing natural language cues\footnote{
Alternatively, learning by active exploration based on generating questions or image captioning~{\citep{mokady2021clipcap}} for the user can be realized~{\citep{Taniguchi2023,activespcoslam2023}.
For example, the robot asks questions to the user, ``{\textit{What kind of place is this?}}''}}, i.e., providing utterances about various locations, such as ``\textit{This is my father Bob's study space, and it has many books}.''
Furthermore, the robot collects multimodal sensor observations from the environment, including images, depth data, odometry, and speech signals. Using these sensor observations, the robot acquires knowledge of the environmental map, connection relationships between places, spatial concepts, and place names.
\\
\textbf{Path planning phase}:
In the planning phase, the robot observes speech instructions such as ``\textit{go to the kitchen}'' as a basic task and ``\textit{go to the kitchen through the bedroom}'' as an advanced task (see Fig.~\ref{fig:overview} (right)).
In particular, this study was focused on hierarchical path planning in advanced tasks.
%
\addspan{
Although the shortest path may not always be the most suitable, robots can select alternative paths to avoid certain areas or perform specific tasks based on the user's instructions. 
For example, the robot might choose a different route to avoid the living room with guests or to check on the pets in the bedroom. 
Thus, users can guide the robot to an improved path by specifying waypoints. 
Furthermore, when multiple locations have the same names (e.g., three bedrooms), selecting the closest route among them is appropriate. 
By specifying the closest waypoint to the target, the robot can accurately select the target even if many places share the same name.
}

\addspan{
In this study, ``optimal'' refers to the scenario that maximizes the probability of a trajectory distribution under given conditions. 
Specifically, the robot should plan an overall optimal path through the designated locations. 
This ensures that the robot's path planning is practical and reduces travel distance and time by considering real-world constraints and objectives. 
It also allows for greater flexibility in guiding the robot through waypoints, enabling users to direct it along preferred routes while maintaining overall effectiveness.
}


This paper proposes a spatial concept-based topometric semantic mapping for hierarchical path planning (SpCoTMHP) with a probabilistic generative model\footnote{The source code is available at \url{https://github.com/a-taniguchi/SpCoTMHP.git}.}.
The topometric semantic map enables path planning that combines abstract place transitions and geometrical structures in the environment.
SpCoTMHP is based on a probabilistic generative model that integrates metric, topological, and semantic levels, as well as speech-and-language models, into a unified framework.
Learning occurs through unsupervised learning of the joint posterior distribution derived from multimodal observations.
To enhance the capture of topological structures, a learning method inspired by the function of replay in the hippocampus is introduced~\citep{Foster2006}.
Ambiguities related to locations and words are addressed through a probabilistic approach informed by robot experience.
In addition, we develop approximate inference methods for effective path planning, where each hierarchy level can influence the others.
The proposed path planning is theoretically supported based on \textit{control as probabilistic inference} (CaI)~\citep{levine2018reinforcement}.
The CaI bridged the theoretical gap between probabilistic inference and control problems, including reinforcement learning. 

The proposed approach is based on \textit{symbol emergence in robotics}~\citep{taniguchi2015symbol,taniguchi2018TCDSsurvey} and provides the advantage of enabling navigation using unique spatial divisions and local names learned without annotations, which are tailored to each individual family or community environment. 
Hence, users can simply communicate with the robot throughout the process, from the learning phase to task execution, thus eliminating the need for robotics expertise.
Moreover, the approach is based on the robot's real-world experience, which enables daily behavioral patterns to be captured, such as where to travel more/less frequently.

We conducted experiments in home environments using the Toyota human support robot (HSR) on the SIGVerse simulator~\citep{inamura2021sigverse} and in a lab-office environment with the real robot Albert~\citep{Radish}.
SpCoTMHP was compared to baseline hierarchical path planning methods in navigation experiments using speech instructions with a designated waypoint.
%
The main contributions of this study are as follows:
\begin{enumerate}[1. ]
    \item We demonstrated that hierarchical path planning, incorporating topological maps through probabilistic inference, achieves \textbf{higher success rates and shorter computation times} for language instructions involving waypoints, compared to methods utilizing heuristic costs.
    \item We illustrated that semantic mapping based on spatial concepts, considering topological maps, achieves \textbf{higher learning performance} than SpCoSLAM, which does not incorporate topological maps.
\end{enumerate}
In particular, the significance of this contribution is characterized by the following four items:
\begin{enumerate}[1. ]
    \item \textbf{\underline{Integrated Learning-Planning Model}}: \textbf{The learning-planning integrated model} could autonomously construct hierarchical spatial representations, including topological place connectivity, from the multimodal observations of the robot, thus \textbf{leading to improved performance in learning and planning}.
    \item \textbf{\underline{Probabilistic Inference for Real-time Planning}}: The approximate \textbf{probabilistic inference based on CaI} enables \textbf{the real-time planning of adaptive paths} from waypoint and goal candidates.
    \item \textbf{\underline{Many-to-Many Relationships for Path Optimization}}: The probabilistic many-to-many relationships between words and locations enable \textbf{planning closer paths when there are multiple target locations}.
    \item \textbf{\underline{Spatial Concepts for Environment-Specific Planning}}: The spatial concepts learned in real environments are effective for \textbf{path planning with environment-specific words}.
\end{enumerate}

The remainder of this paper is organized as follows:
Section~\ref{sec:related_work} presents related work on topometric semantic mapping, hierarchical path planning, and the spatial concept-based approach.
Section~\ref{sec:proposed_method} describes the proposed method SpCoTHMP.
Section~\ref{sec:exp1} presents experiments performed using a simulator in multiple home environments.
Section~\ref{sec:exp2} discusses experiments performed in real environments.
Finally, Section~\ref{sec:conclusion} concludes the paper.

\section{Related Work}
\label{sec:related_work}

This section describes the topometric semantic mapping in Section~\ref{sec:related_work:semantic_map}, the hierarchical path planning in Section~\ref{sec:related_work:planning}, robotic planning using large language models (LLMs) and foundation models in Section~\ref{sec:related_work:llm}, and the spatial concept-based approach in Section~\ref{sec:preliminary_spco}.
Table~\ref{tbl:related_work_map} displays the main characteristics of the map representation and the differences between related works.
Table~\ref{tbl:related_work_hpp} presents the main characteristics of path planning and the differences between related works.

\subsection{Topometric semantic mapping}
\label{sec:related_work:semantic_map}
\begin{table}[tb]
    \begin{center}
    \caption{
        The main characteristics of map representation and the difference between related works.
    }
    \begin{tabular}{lcccp{146pt}} \hline
    \textbf{References}         & Metric     & Topological  & Semantic   & Class-label / Vocabulary    \\ \hline
    \citet{Shatkay2002}         & \checkmark & \checkmark   & ---        & ---           \\
    \citet{rangel2016lextomap}  & \checkmark & \checkmark   & \checkmark & Preset label  \\
    \citet{Zheng2018}           & \checkmark & \checkmark   & \checkmark & Preset label  \\
    \citet{Karaoguz2016}        & \checkmark & ---          & \checkmark & Preset label  \\
    \citet{Kostavelis2016}      & \checkmark & \checkmark   & \checkmark & Preset label  \\
    \citet{Luperto2018}         & \checkmark & ---          & \checkmark & Preset label  \\
    \citet{Gomez2020}           & \checkmark & \checkmark   & \checkmark & Free area or transit area (door) \\
    \citet{Rosinol2021}         & \checkmark & \checkmark   & \checkmark & Preset label  \\
    \citet{Hiller2019}          & \checkmark & \checkmark   & \checkmark & Preset label  \\
    \citet{Sousa2021}           & \checkmark & ---          & \checkmark & Preset label  \\
    \citet{ataniguchi_IROS2017,ataniguchi2020spcoslam2}  & \checkmark & --- & \checkmark & On-site learning (environment-specific words) \\
    SpCoTMHP (Ours)             & \checkmark & \checkmark   & \checkmark & On-site learning (environment-specific words) \\
    \hline 
    \end{tabular}
    \label{tbl:related_work_map}
    \end{center}
\end{table}

For bridging the topological-geometrical gap, geometrically-constrained hidden Markov models, as a method based on probabilistic models, were proposed for robot navigation in the past~\citep{Shatkay2002}.
The similarity between these models and that proposed in this study is that probabilistic inference realized path planning.
However, those earlier models did not introduce semantics, such as location names.

Research on semantic mapping has been increasingly emphasized in recent years. 
In particular, semantic mapping assigns place meanings to the map of a robot~\citep{kostavelis2015semantic, Garg2020}.
However, numerous existing studies provide a preset location label for an area on a map.
For example, LexToMap~\citep{rangel2016lextomap} assigns convolutional neural network-recognized lexical labels to a topological map.
The proposed approach enables unsupervised learning based on multimodal perceptual information for categorizing unknown places and flexible vocabulary assignments.

The use of topological structures enables more accurate semantic mapping~\citep{Zheng2018}.
The proposed method is expected to improve its performance by introducing topological levels.
The nodes in a topological map can vary depending on the methods, such as room units or small regions~\citep{Karaoguz2016,Kostavelis2016,Luperto2018,Gomez2020}.
Kimera~\citep{Rosinol2021} used multiple levels of spatial hierarchical representation, such as metrics, rooms, places, semantic levels, objects, and agents. 
In this study, the robot automatically determined the spatial segmentation unit based on experience.

In several semantic mapping studies~\citep{Hiller2019,Sousa2021}, topological semantic maps were constructed from visual images or metric maps using convolutional neural networks. 
However, these studies did not consider path planning.
In contrast, the proposed method is characterized by an integrated model that includes learning and planning.

\subsection{Hierarchical path planning}
\label{sec:related_work:planning}

\begin{table}[tb]
    \begin{center}
    \caption{
        The main characteristics of path planning and the difference between related works. 
    }
    \begin{tabular}{p{120pt}p{100pt}p{130pt}p{90pt}} \hline
    \textbf{References}         & Planning approach         & Instruction for navigation & Goal determination  \\ \hline
    \citet{Holte1996}           & Classical (A$^\star$)     & ---         & Explicitly given as a point \\
    \citet{Kostavelis2016}      & Dijkstra and LSTM         & \textit{go-to} commands by GUI  & Explicitly given by the user \\
    \citet{Stein2020}           & Learned Subgoal Planning & ---         & Explicitly given as a point \\ 
    \citet{Rosinol2021}         & Multi-level A$^\star$ & Semantic queries & Explicitly given from queries \\
    \citet{Kulkarni2016a,Haarnoja2018a}       & Hierarchical reinforcement learning & ---         & Autonomously estimated \\
    \citet{Krantz2020,Gu2022,Huang2023}          & VLN & Unambiguous and detailed description  &  Non-Explicit (Vision-based)  \\
    \citet{anderson2018vision,Chen2020a}  & Deep reinforcement learning & (Same as above) & (Same as above)  \\
    \citet{ataniguchi2020spconavi} & Cal framework & Daily short speech sentence (containing environment-specific words) & Non-Explicit (Probabilistic)  \\
    SpCoTMHP (Ours)             & Hierarchical Cal framework & Daily short speech sentence (containing environment-specific words and waypoints) & (Same as above) \\
    \hline 
    \end{tabular}
    \label{tbl:related_work_hpp}
    \end{center}
\end{table}
%

Hierarchical path planning has been a significant topic of study for a long period, e.g., hierarchical A$^\star$~\citep{Holte1996}.
Using topological maps for path planning (including learning paths between edges) is more effective in reducing computational complexity than considering only the movement between cells in a metric map~\citep{Kostavelis2016,Stein2020,Rosinol2021}. 
In addition, the extension of map representations to hierarchical semantic maps enables navigation based on speech.

Given that the proposed method realizes a hierarchy based on the CaI framework~\citep{levine2018reinforcement}, it is theoretically connected with hierarchical reinforcement learning.
In hierarchical reinforcement learning, sub-goals and policies are autonomously estimated \citep{Kulkarni2016a,Haarnoja2018a}.
This study investigated tasks similar to hierarchical reinforcement learning to infer probabilistic models.
As models for probabilistic inference, it is expected to be theoretically readable and integrated with other methods.

Vision-and-language navigation (VLN) aims to help an agent navigate through an environment based on natural language instructions while using visual information from the environment~\citep{Krantz2020,Gu2022,Huang2023}.
This study differs from those on VLNs in several respects.
The first is the complexity of the instructions.
In VLN tasks, unambiguous and detailed natural language instructions are given.
In contrast, the proposed method engaged in tasks characterized by the degree of shortness and ambiguity with which people speak daily.
The second was the training scenario.
The VLN dataset uses only common words annotated by people in advance.
In contrast, the proposed approach can address spatial words in communities living in specific environments.
Third, although VLNs use vision during path planning, vision was used to generalize spatial concepts only during training of the proposed method.
This is due to the difference between sequential action decisions and global path planning.
Finally, in recent studies on VLN, deep and reinforcement learning techniques were used~\citep{anderson2018vision, Chen2020a}. 
The proposed probabilistic model autonomously navigated toward the target location using speech instructions as a modality.

\subsection{Robotic planning using LLM and foundation models}
\label{sec:related_work:llm}

Recently, there has been a growing utilization of LLMs and foundational models to enhance robot autonomy ~\citep{Vemprala2023chatgpt_robotics,Firoozi2023,Zeng2023LLMforRobotics}.
SayCan~\citep{Ahn2022palm_saycan} integrates pre-trained LLMs and behavioral skills to empower robots to execute context-aware and appropriate actions in real-world settings.
In this approach, the LLM conducts higher-level planning based on language, while facilitating lower-level action decisions grounded in physical constraints.
However, a key challenge remains in accurately capturing the characteristics of physical space, such as the presence of walls, distance, and room shape, solely using LLMs.
In contrast, our study tightly integrates language, spatial semantics, and physical space to estimate trajectories comprehensively.
Furthermore, our proposed method is designed to complement LLM-based planning and natural language processing, with the expectation of seamless integration.

Several studies employ LLMs and foundational models to accomplish navigation tasks. 
LM-Nav~\citep{Shah2022LM-Nav} has integrated contrastive language-image pretraining (CLIP)~\citep{Radford2021clip} and generative pretrained transformer-3 (GPT-3)~\citep{Brown2020gpt3}. This system enables navigation directly from language instructions and robot-perspective images alone. However, this approach necessitates a substantial amount of driving data from the target environment.

Conversely, an approach that combines vision-language models (VLMs) and semantic maps has also appeared.
CLIP-Fields~\citep{MahiShafiullah2023}, natural language maps (NLMap)~\citep{Chen2023NLMap}, and VLMaps~\citep{Huang2023} use LLMs and VLMs to create 2D or 3D space and language associations to enable navigation for natural language queries.
These mainly record the placement of objects on the map, and cannot understand the meaning of locations and planning for each location.
Additionally, LLM/VLM-based approaches have a large common sense vocabulary, similar to an open vocabulary. 
\addspan{
However, using pre-trained place recognizers alone makes it difficult to handle environment-specific names (e.g., Alice's room). 
While LLMs have the potential to handle environment-specific names through in-context learning, they are not yet integrated with mapping and navigation in existing models. 
}
Our spatial concept-based approach can address knowledge specific to the home environment through on-site learning.

\subsection{{Spatial concept-based approach}}
\label{sec:preliminary_spco}

In Section~\ref{sec:proposed_method}, we present two major previous studies on which the proposed method is based.

As presented in previous research, SpCoSLAM~\citep{ataniguchi_IROS2017,ataniguchi2020spcoslam2} forms spatial concept-based semantic maps based on multimodal observations obtained from the environment.
Here, multimodal observations for spatial concept formation refer to image, depth sensor values, odometry, and speech signals.
Moreover, it can acquire novel place categories and vocabularies from unknown environments.
However, SpCoSLAM cannot estimate the topological level, i.e., whether one place is spatially connected to another.
The details of the formulation of the probabilistic generative model are described in Appendix~\ref{apdx:SpCoSLAM:overview:formulation}.
The learning procedure for each step is described in Appendix~\ref{apdx:SpCoSLAM:learning:procedure}.
In this study, we applied the hidden semi-Markov model (HSMM)~\citep{Johnson2013}, which estimates the transition probabilities between places and constructs a topological graph, instead of the Gaussian mixture model (GMM) part in SpCoSLAM.
%

In addition, SpCoNavi~\citep{ataniguchi2020spconavi} plans the path in the CaI framework~\citep{levine2018reinforcement}, which focuses on the action decision in the probabilistic generative model of SpCoSLAM.
The detail of the formulation of CaI is described in Appendix~\ref{apdx:SpCoNavi:CaI}. 
Notably, SpCoNavi realized navigation from simple speech instructions using a spatial concept acquired autonomously by the robot. 
However, SpCoNavi does not demonstrate hierarchical path planning, and scenarios specifying a waypoint are not considered.
In addition, there are several problems to be solved in that 
SpCoNavi, which is based on the Viterbi algorithm~\citep{viterbi1967error}, is computationally expensive, given that all grids of the occupied grid map are used as the state space. Moreover, it is vulnerable to the real-time performance required for robot navigation.
Furthermore, SpCoNavi, which is based on the A$^\star$ approximation, reduces the computational cost; however, its performance is inferior to that of the Viterbi.
Therefore, in this study, we utilized a topological semantic map based on spatial concepts to reduce the number of states and rapidly infer possible paths between each state.

\section{Proposed Method: SpCoTMHP}
\label{sec:proposed_method}

This paper proposes a spatial concept-based topometric semantic mapping for hierarchical path planning (SpCoTMHP).
Spatial concepts refer to categorical knowledge of places from multimodal information through unsupervised learning.
The proposed method realizes efficient navigation from human speech instructions through inference based on a probabilistic generative model.
The proposed approach enhances human comprehensibility and explainability for communication by employing Gaussian distributions as the fundamental spatial units (i.e., representing a single place).
The capabilities of the proposed generative model are as follows:
(i) Place categorization by extracting connection relations between places using unsupervised learning,
(ii) Many-to-many correspondence between words and places,
and 
(iii) Efficient hierarchical path planning by introducing two variables ($t$ and $e$) with different time constants.

In probabilistic generative models, three phases can be distinguished: (a) the model definition in the probability distribution of the generative process (shown in Section \ref{sec:SpCoSLAM-TM:generative_model}), (b) inference of the posterior distribution for parameter learning (shown in Section \ref{sec:SpCoSLAM-TM:learning}), and (c) probabilistic inference for task execution after learning (shown in Sections \ref{sec:SpCoNavi-HP:CaI} and \ref{sec:SpCoNavi-HP:approx}).

\subsection{Definition of probabilistic generative model}
\label{sec:SpCoSLAM-TM:generative_model}

\begin{figure*}[!tb]
    \begin{center}
    \includegraphics[width=0.96\linewidth, clip]{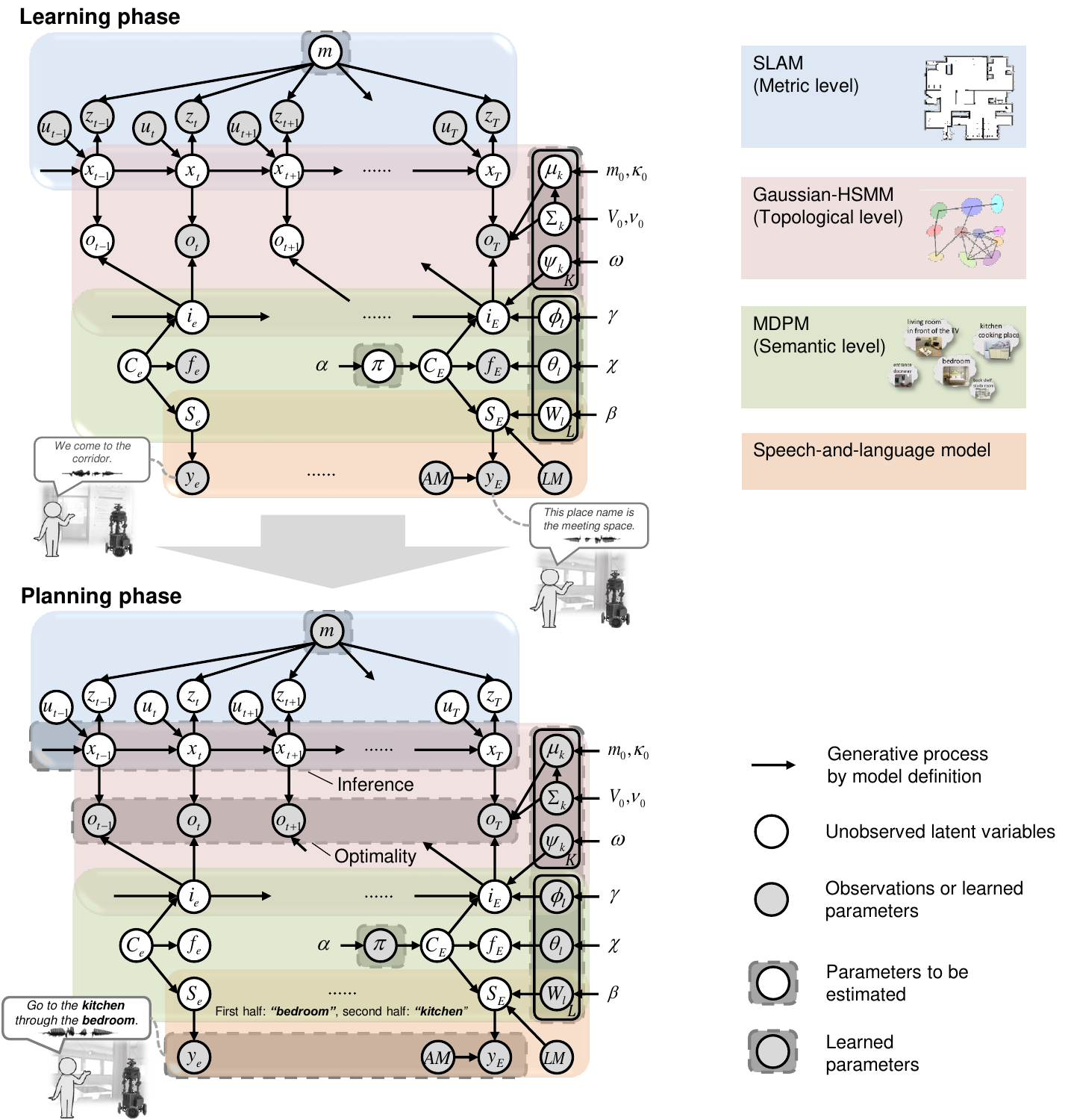} 
    \caption{
        Graphical model representation of SpCoTMHP (top) spatial concept learning and path planning phases (bottom).
        The two phases imply different probabilistic inferences for the same generative model.
        This has the mathematical advantage that different probabilistic inferences can be applied under the same model assumptions.
        The integration of several parts as a single model allows inferences to consider various probabilities throughout.
        The graphical model represents the conditional dependency between random variables.
        Gray nodes indicate observations or learned parameters as fixed conditional variables. 
        White nodes denote unobserved latent variables to be estimated.
        Arrows from the global variables to local variables other than $T$ and $E$ are omitted.
        In the learning phase, multimodal observations are obtained several times.
        Based on these observables, latent variables are estimated.
        In the planning phase, the parameters estimated in the learning phase and optimality variables are given.
        Under these conditions, the distribution of trajectories is estimated.
        \addspan{$D_{e}$ was omitted from the graphical model representation.}
    }
    \label{fig:graphical-model}
    \end{center}
\end{figure*}

\begin{table}[tb]
    \begin{center}
    \caption{Description of the random variables used in the proposed model}
    \begin{tabular}{cp{340pt}} \hline
        \textbf{Symbol} & \textbf{Definition}\\ \hline
        $m$     & Environmental map (occupancy grid map)\\ 
        $x_t$   & Self-position of the robot (state variable) \\ 
        $u_t$   & Control data (action variable) \\ 
        $z_t$   & Depth sensor data \\ 
        $o_{t}$ & Optimality variable (event-driven) \\
        $D_{e}$ & Duration length for $o_{t}$ in $i_{e}$ \\
        $i_{e} \in \{ k \} $ & Category index of the position distributions \\ 
        $C_{e} \in \{ l \} $ & Category index of spatial concepts \\         
        $f_{e}$ & Visual features of the camera image \\ 
        $y_{e}$ & Speech signal of the uttered sentence \\ 
        $S_{e}$ & Word sequence in the uttered sentence \\ 
        {$\mu_{k}$, $\Sigma_{k}$} & {Parameters of multivariate Gaussian distribution \,\, (position distribution)} \\ 
        {$\psi_{k}$}   & {Parameter of state-transition for $i_{e}$ in $i_{e-1}= k$} \\ 
        {$\pi $}       & {Parameter of mixture weights for $C_{e}$} \\ 
        {$\phi_{l}$}   & {Parameter of mixture weights for $i_{e}$ in $C_{e}=l$} \\ 
        {$\theta_{l}$} & {Parameter of feature distribution for $f_{e}$}  \\ 
        {$W_{l}$}      & {Parameter of word distribution for $S_{e}$} \\ 
        $LM$ & Language model (n-gram and word dictionary)\\  %
        $AM$ & Acoustic model for speech recognition \\ 
        $\alpha $, $\beta$, $\gamma$, $\chi$, $\omega$, & {Hyperparameters of prior distributions} \\
        $m_{0}$, $\kappa_{0}$, $V_{0}$, $\nu_{0}$& \\ \hline
        $T$  & Final time of robot operation \\
        $E$  & Total number of user utterances (in the learning phase) or total number of location moves (in the planning phase) \\
        $L$  & Total number of spatial concepts \\
        $K$  & Total number of position distributions \\ 
         \hline
    \end{tabular}
    \label{tab:graphical-model}
    \end{center}
\end{table}

SpCoTMHP is designed as an integrated model for each module: SLAM, HSMM, multimodal Dirichlet process mixture~(MDPM) for place categorization, and the speech-and-language model.
Therefore, it is simple to distribute development and further the module coupling in the framework of Neuro-SERKET~\citep{taniguchi2020neuroSERKET}.
The integrated model has the advantage that the inference functions as a whole, complementing each uncertainty.
Figure~\ref{fig:graphical-model} presents the graphical model representation of SpCoTMHP, and Table~\ref{tab:graphical-model} lists each variable of the graphical model. 
Unlike SpCoSLAM~\citep{ataniguchi_IROS2017}, SpCoTMHP introduces two different time units (real-time robot motion-based time step $t$ and event-driven time step $e$) and extends GMM to HSMM.
The events represent the timing of the user utterances during the learning and the switching of locations visited during planning.
The generative process (prior distribution or likelihood function) is defined by the graphical model representation of SpCoTMHP.

\textbf{SLAM part (metric level):} 
The probabilistic generative model of SLAM can represent the time-series transition of self-position.
The state space on the map corresponds to the metric level.
These probability distributions were standard in SLAM for probabilistic approaches~\citep{thrun2005probabilistic}.
Accordingly, Eq.~(\ref{eq:seisei10}) represents a measurement model that is a likelihood of a depth sensor $z_{t}$ at a given position $x_{t}$ and map $m$.
Eq.~(\ref{eq:seisei9}) represents a motion model that is a state-transition related to the position $x_{t}$ based on action $u_{t}$ in previous position $x_{t-1}$ in SLAM as follows:
\begin{flalign}
    z_{t}      &\sim p(z_{t} \mid x_{t}, m), \label{eq:seisei10} & t &= 1, 2, \dots, T \\ 
    x_{t}      &\sim p(x_{t} \mid x_{t-1}, u_{t}). \label{eq:seisei9} 
\end{flalign}
Self-localization assumes a transition at time $t$ due to the motion of the robot.
The variable $x_t$ is shared with the HSMM.

\textbf{HSMM part (from metric level to topological level):} 
The HSMM can cluster each location data of the robot as position distributions and represent probabilistic transitions between position distributions.
This refers to transitioning from the metric level to the topological level.
The HSMM connects two units: time $t$ and event $e$.
A binary random variable that indicates whether there is an event is defined as follows:
\addspan{{
\begin{flalign}
    o_{t} &\sim p(o_{t} \mid x_{t}, i_{e}, \mu , \Sigma;  \{ D_{e} \}) = 
    \begin{cases}
        \eta \cdot {\mathcal N}(x_{t} \mid \mu _{i_{e}}, \Sigma_{i_{e}}) & \text{if } o_{t} = 1, \\
        1 - \eta \cdot {\mathcal N}(x_{t} \mid \mu _{i_{e}}, \Sigma_{i_{e}}) & \text{if } o_{t} = 0,
    \end{cases} & t &= t_{e}, \dots, t'_{e} 
    \label{eq:ot}
\end{flalign}
}}
where \addspan{$\eta = 1 / \sum_{j = 1}^{K} \mathcal{N}(x_{t} \mid \mu_{j}, \Sigma_{j})$ is the normalization constant,} ${\mathcal N}()$ is a multivariate Gaussian distribution, $\mu = \{ \mu _{k} \}$, $\Sigma = \{ \Sigma_{k} \}$, and $o_{t}=1$ denotes that the event occurred at time~$t$.
Here, $o_{t} \in \{ 0, 1 \}$ takes a binary value.
This event-driven variable corresponds to the optimality variable in the CaI~\citep{levine2018reinforcement}.
The duration distribution assumes uniform distribution in $[1,T]$: 
\begin{flalign}
    D_{e}      &\sim {\rm Unif}(1,T), & e &= 1, 2, \dots, E 
    \label{eq:dt}
\end{flalign}
where the equation relating $t$ and $e$ is 
$t_{e} = \sum_{e' < e} D_{e'}$,
and 
the final time at the event $e$ is
$t'_{e} = t_{e} + D_{e} - 1$.
Thus, $E \leqq T$ and $T = \sum_{e=1}^{E} D_{e}$.
\\
The position distribution represents a coherent unit of place and is represented by a Gaussian distribution, i.e., as a node in a topological map.
$\mu_k$ represents the representative point of node $k$ on the map.
$\Sigma_k$ represents the spread of the place of node $k$.
To capture transitions between places, as connection weights between nodes to represent edges in the topological map, $\psi_{k}$ is introduced as follows:
\begin{flalign}
    \mu_{k}    &\sim {\mathcal N}( m_{0}, \Sigma_{k} / \kappa _{0} ), \label{eq:seisei8} & k &= 1, 2, \dots, \infty \\    
    \Sigma_{k} &\sim {\mathcal IW}( V_{0}, \nu _{0} ),  \label{eq:seisei7} \\ 
    \psi_{k}   &\sim {\rm DP}(\omega ), \label{eq:seisei5-2} 
\end{flalign}
where ${\mathcal IW()}$ is the inverse-Wishart distribution, and ${\rm DP()}$ represents the Dirichlet process.
{The Dirichlet process assumes an infinite number of categories, which allows for infinite mixed HSMM, thereby enabling the learning of positional distributions, i.e., nodes of a topological map, that flexibly depend on the environment.}
The inverse-Wishart distribution is a conjugate prior distribution on the covariance matrix of the Gaussian distribution. 
The conjugate prior distribution was established because it provides the advantage that the posterior distribution can be obtained analytically.
Please refer to the literature on machine learning~\citep{murphy2012machine} for the specific formulas of these probability distributions.

\textbf{HSMM + MDPM connection (from topological level to semantic level):}
The variable $i_{e}$ of the topological node is shared between HSMM and MDPM.
The probability distribution of $i_{e}$ for connecting two modules is defined by unigram rescaling (UR)~\citep{gildea1999topic} as follows: 
\begin{flalign}
    i_{e}      &\sim p(i_{e} \mid {i_{e-1}}, \psi, {C_{e}}, \phi ) \label{eq:seisei6} \\ 
    &\quad \overset{\mathrm{UR}}{ \approx } \underbrace{{\rm Mult}(i_{e} \mid \psi_{i_{e-1}})}_{\substack{\text{Transition prob.} \\ \text{by HSMM}}} \underbrace{\frac{{\rm Mult}(i_{e} \mid \phi_{C_{e}})}{\sum_{c'=1}^{L} {\rm Mult}(i_{e} \mid \phi_{c'})}}_{\substack{\text{Category dependent term} \\ \text{/ Rescaling term}}},
    \label{eq2:ie_UR}
\end{flalign}
where $\psi = \{ \psi _{k} \}$, $\phi = \{ \phi _{l} \}$, and ${\rm Mult()}$ is a multinomial distribution.
The first term in Eq.~(\ref{eq2:ie_UR}) denotes the transferability between places, and the second term denotes the correspondence between the spatial concept and the position distribution.
The position distribution $k=i_e$ takes high probability when it corresponds to the spatial concept $C_e$ and is connected to the position distribution $i_{e-1}$.

\textbf{MDPM part (semantic level):}
The MDPM is a mixture distribution model for forming place categories from multimodal observations.
Through the spatial concept $l=C_e$, the probabilities of the modalities represented by $\phi_{l}$, \addspan{{$\theta_{l}$}}, and $W_{l}$ are corresponded.
The MDPM is positioned at the semantic level, which represents spatial concepts based on places $i_{e}$, speech-language $S_{e}$, and image features $f_{e}$ as follows:
\begin{flalign}
    \pi        &\sim {\rm DP}(\alpha ), \label{eq:seisei1} \\
    \phi_{l}   &\sim {\rm DP}(\gamma ), \label{eq:seisei5} & l &= 1, 2, \dots, \infty \\
    \addspan{\theta_{l}} &\sim {\rm Dir}(\chi ), \label{eq3:seisei3} \\ 
    W_{l}      &\sim {\rm Dir}(\beta), \label{eq:seisei3}  \\ 
    C_{e}      &\sim {\rm Mult}(\pi), \label{eq:seisei2} & e &= 1, 2, \dots, E \\
    f_{e}      &\sim {\rm Mult}(\theta_{C_{e}}), \label{eq3:seisei2}  
\end{flalign}
where ${\rm Dir()}$ is the Dirichlet distribution.
According to the data, the DP automatically determines the numbers of spatial concepts $L$ and position distributions $K$.
A multinomial distribution is applied to the discrete variables. The Dirichlet distribution and DP are set up as conjugate prior distributions for the multinomial distribution.

\textbf{MDPM + language model connection (semantic level):}
Variable of word sequence $S_{e}$ is shared between the MDPM and the language model.
The probability distribution of $S_{e}$ for connecting two modules is defined by UR~\citep{gildea1999topic} as follows: 
\begin{flalign}
    S_{e}      &\sim p(S_{e} \mid C_{e}, W, LM) \label{eq2:seisei4b} \\  
    &\quad \overset{\mathrm{UR}}{ \approx } \underbrace{p(S_{e} \mid LM)}_{\text{N-gram prob.}} \prod_{b=1}^{B_{e}} \underbrace{\frac{{\rm Mult}(S_{e,b} \mid W_{C_{e}})}{\sum_{c'=1}^{L} {\rm Mult}(S_{e,b} \mid W_{c'})}}_{\substack{\text{Category dependent term} \\ \text{/ Rescaling term}}},
    \label{eq2:st_UR}    
\end{flalign}
where $W = \{ W_{l} \}$.
Moreover, $B_{e}$ is the number of words in the sentence, and $S_{e,b}$ is the $b$-th word in the sentence at event~$e$.
The first term in Eq.~(\ref{eq2:st_UR}) is the probability of occurrence of a word based on the n-gram language model $LM$.
Specifically,
$p(S_{e} \mid LM) = \prod_{b=1}^{B_{e}} p(S_{e,b} \mid S_{e,b-n+1:b-1}; LM)$.
The second term is the spatial concept-dependent word probability distribution, which is computed for each word independently.

\textbf{Speech-and-language model part:}
The generative process as the likelihood of speech given a word sequence is as follows:
\begin{flalign}
    y_{e}      &\sim p(y_{e} \mid S_{e}, AM). \label{eq2:seiseiYt} 
\end{flalign}
This probability distribution does not usually appear explicitly but is internalized as an acoustic model in probability-based speech recognition systems.

\subsection{Spatial concept learning as topometric semantic mapping}
\label{sec:SpCoSLAM-TM:learning}

The joint posterior distribution is described as
\begin{align}
    p(x_{0:T},S_{1:E},\mathbf{C}_{1:E}, \mathbf{\Theta} %
    \mid u_{1:T}, z_{1:T}, o^{\ast}_{t'_{1:E}}, y_{1:E}, f_{1:E} ,\mathbf{h}), 
    \label{eq:SpCoTMHP}
\end{align}
where the set of latent variables is denoted by $\mathbf{C}_{1:E} = \{i_{1:E},C_{1:E} \}$, the set of global model parameters $\mathbf{\Theta} = \{ m, \mu, \Sigma, \psi, \pi, \phi, \theta, W, LM, AM \}$, and the set of hyperparameters $\mathbf{h}= \{ \alpha,\beta,\gamma,\chi,\omega, m_{0},\kappa_{0}, V_{0},\nu_{0} \}$.
The set of event-driven variables is $o^{\ast}_{t'_{1:E}} = \{ o_{t'_{e}}=1 \}^{E}_{e=1}$.
 
In this paper,
as an approximation to sampling from Eq.~(\ref{eq:SpCoTMHP}), the parameters are estimated as follows:
\begin{align}
    x_{0:T}, m &\sim p(x_{0:T}, m \mid u_{1:T}, z_{1:T}), \label{eq:slam} \\
    S_{e} &\sim p(S_{e} \mid y_{e}, LM, AM), \quad e = 1, 2, \dots, E \label{eq:speech_recog}  \\
    \mathbf{C}_{1:E}, \mathbf{\Theta}^{\prime} 
    &\sim p(\mathbf{C}_{1:E}, \mathbf{\Theta}^{\prime} %
    \mid x_{0:T}, o^{\ast}_{t'_{1:E}}, S_{1:E}, f_{1:E} ,\mathbf{h}),  \label{eq:spco_gibbs} 
\end{align}
where $\mathbf{\Theta}^{\prime} = \{ \mu, \Sigma, \psi, \pi, \phi, \theta, W \}$.
Equation~(\ref{eq:slam}) is realized by grid-based FastSLAM 2.0~\citep{gridbasedfastslam2007}.
Equation~(\ref{eq:speech_recog}) represents the speech recognition of $y_{e}$. $LM$ and $AM$ were pre-set.
The proposed method can handle uncertainty in speech recognition by capturing the $N$-best speech recognition results as a Monte Carlo approximation.
The variables in Eq.~(\ref{eq:spco_gibbs}) can be learned using Gibbs sampling, which is a Markov chain Monte-Carlo-based batch learning algorithm, specifically, the weak-limit and direct-assignment sampler~\citep{Johnson2013}.

In the learning phase, the user provides a teaching utterance each time the robot transitions between locations.
Given that the utterance is event-driven, it was assumed that the variables on the spatial concepts are observed only at event $e$.
Here, the time of the $e$-th event (when the robot observes that an utterance indicates the place) is $t'_{e}$.
In particular, $o_{t'_{e}}=1$ is observed at the instants of $t'_{e}$, and $o_{t}$ is unobserved at other times.
Therefore, the inference for learning $i_{e}$ is equivalent to an HMM.

\textbf{Reverse replay:}
In the case of spatial movement, we can transition from $i_{e-1}$ to $i_{e}$, or vice versa.
Therefore, $i'_{E:1}$, which is replayed using the steps of $e$ in reverse order, can be used for learning when sampling $\psi$.
This is based on the replay performed in the hippocampus of the brain~\citep{Foster2006}.

\subsection{Hierarchical path planning by control as inference}
\label{sec:SpCoNavi-HP:CaI}

The probabilistic distribution, which represents trajectory $\tau=\{ u_{1:T}, x_{1:T} \}$ when a speech instruction $y_{e}$ is given, is maximized to estimate an action sequence $u_{1:T}$ (and the path $x_{1:T}$ on the map) as follows:
\begin{eqnarray}
    u_{1:T} &=& \argmax_{u_{1:T}} p(\tau \mid o^{\ast}_{1:T}, y_{1:E}, x_{0}, \mathbf{\Theta}). 
    \label{eq:spconavi-HP}
\end{eqnarray}
The planning horizon at metric level $T$ is the final time of the entire task when a one-time step traverses one grid block on the metric map.
The planning horizon at the topological level $E$ is the number of event steps used to navigate by speech instruction.
As shown in Eqs. (\ref{eq:ot}) and (\ref{eq:dt}), each event step $e$ corresponds to time series ${t_{e}:t'_{e}}$.
The metric-level planning horizon in Step $e$ corresponds to the duration $D_{e}$ of the HSMM.
In the metric-level planning horizon, the event-driven variable is always $o^{\ast}_{1:T} = \{ o_{t}=1 \}_{t=1}^{T} $ by the CaI.
Speech instruction $y_{e}$ is assumed to be the same as that from $e=1$ to $E$.
This indicates that $o_{t}$ and $y_{e}$ are multiple optimalities in terms of CaI~\citep{kinose2020integration}.
From the above, Eq.~(\ref{eq:spconavi-HP}) is as follows:
\begin{align}
    & p(\tau \mid o^{\ast}_{1:T}, y_{1:E}, x_{0}, \mathbf{\Theta})  \nonumber \\
    & \approx \prod_{e=1}^{E} \left[ \sum_{i_{e}=1}^{K} \frac{ {\rm Mult}(i_{e} \mid  \psi_{i_{e-1}}) }{ \sum_{c'=1}^{L} {\rm Mult}(i_{e} \mid \phi_{c'}) }   \right.  \nonumber \\ 
    &\quad \sum_{{C}_{e}=1}^{L} {\rm Mult}(i_{e} \mid \phi_{C_{e}}) {\rm Mult}(S_{e} \mid W_{C_{e}}) {\rm Mult}(C_{e} \mid \pi)  \nonumber \\ 
    &\quad  \left.  \prod_{t=t_{e}}^{t'_{e}} {{\mathcal N}(x_{t} \mid \mu _{i_{e}}, \Sigma_{i_{e}})}p(x_{t} \mid m) p(x_{t} \mid x_{t-1}, u_{e}) \right], 
    \label{eq:spconavi-HP_detail3}    
    \\ 
    &S_{e} \sim p(S_{e} \mid y_{e}, LM, AM),
\end{align}
where $p(x_{t} \mid m)$ is a probabilistic representation of the cost map, and the maximum limit value of $D_{1:E}$ is given.
In addition, the word sequence $S_{e}$ is obtained by the speech recognition of $y_{e}$ as the bag-of-words of $N$-best.
The assumptions, for example, the SLAM models and cost map, in the derivation of the equation, are the same as those in the SpCoNavi paper~\citep{ataniguchi2020spconavi}.

In this study, we assumed that the robot could extract words that indicate the goal and waypoint from a particular utterance sentence.
In topological-level planning, including the waypoint, the waypoint word is inputted in the first half and the target word in the second half.

\subsection{Approximate inference for hierarchical path planning}
\label{sec:SpCoNavi-HP:approx}

The strict inference of Eq.~(\ref{eq:spconavi-HP_detail3}) requires a double-forward backward calculation.
In this case, reducing the calculation cost is necessary to accelerate path planning, which is one of the objectives of this study.
Therefore, this paper proposes an algorithm to solve Eq.~(\ref{eq:spconavi-HP_detail3}).
Algorithm~\ref{alg:planning} presents the hierarchical planning algorithm produced by SpCoTMHP.

Path planning is divided into topological and metric levels, and the CaI is solved at each level.
Metric-level planning assumes that the partial paths in each transition between places are solved in A$^{\star}$.
The partial paths can be pre-computed regardless of the speech instruction.
Topological-level planning is approximated concerning the probability distribution of $i_{e}$, assuming Markov transitions.
Finally, the partial paths in each transition between places are integrated as a whole path.
Metric and topological planning can influence each other.

\begin{algorithm}[tb] 
    \caption{
        Hierarchical path planning algorithm.
    } 
    \label{alg:planning}
    \begin{algorithmic}[1]
        \State{// \textbf{Pre-calculation}:}
        \State{$\{ \hat{x}_{t'_{e}|i_{e}} \} \sim \mathtt{Gaussian\_Mixture}( \mathbf{\phi} , \mathbf{\mu} , \mathbf{\Sigma} )$}
        \State{Create a graph between waypoint candidates}
        \For{all nodes, $n_{i_{e-1}} \rightarrow n_{i_{e}}$,}
            \State{$\hat{\mathbf{x}}_{i_{e-1},i_{e}}^{[ n_{i_{e-1}}, n_{i_{e}} ]}  \leftarrow {\rm A}^{\star}(\hat{x}_{t'_{e-1}|i_{e-1}}^{[n_{i_{e-1}}]}, \hat{x}_{t'_{e}|i_{e}}^{[n_{i_{e}}]}, w_{e})$}
            \State{Calculate likelihood $\hat{\mathbf{w}}_{i_{e-1},i_{e}}^{[ n_{i_{e-1}}, n_{i_{e}} ]}$ for partial paths}
        \EndFor
        \State{// \textbf{When a speech instruction $y_{e}$ is given}:}
        \State{$S_{e} \leftarrow \mathtt{Speech\_Recognition}(y_{e}, LM, AM)$}
        \State{Estimate an index $i_{0}$ of the place in initial position $x_{0}$}     
        \State{$\mathbf{n}_{1:E}, i_{1:E} \leftarrow \mathtt{Search}(i_{0}, S_{e}, \hat{\mathbf{w}}, \mathbf{\Theta})$}
        \Comment{Eq.~(\ref{eq:spconavi-HP:jyoui2})}
        \State{Connect the partial paths $\mathbf{n}_{1:E}$ as the whole path $\mathbf{x}_{1:E}$}
        \State{$\mathbf{x}_{1:E} \leftarrow \mathtt{Path\_Smoothing}(\mathbf{x}_{1:E}, m)$}
        \Comment{optional process}
    \end{algorithmic}
\end{algorithm}

Path planning at the metric level (i.e., partial path $\mathbf{x}_{i_{e-1},i_{e}}$ when transitioning from $i_{e-1}$ to $i_{e}$) is described as follows:
\begin{align}
    x_{t_{e}:t'_{e}}
    &= \argmax_{x_{t_{e}:t'_{e}}} 
    \prod_{t=t_{e}}^{t'_{e}} {{\mathcal N}(x_{t} \mid \mu_{i_{e}}, \Sigma_{i_{e}})} \nonumber \\
    &\qquad \qquad \qquad p(x_{t} \mid m) p(x_{t} \mid x_{t-1}, u_{t}). 
    \label{eq:spconavi-HP:kai}
\end{align}
This indicates that the inference of a metric-level path can be expressed in terms of CaI.

Calculating Eq.~(\ref{eq:spconavi-HP_detail3}) for all possible positions was difficult.
Therefore, we used the mean or sampled values from the Gaussian mixture of position distributions as a goal position candidate, i.e., $\hat{x}_{t'_{e}|i_{e}}^{[n_{i_{e}}]} \sim {\mathcal N}(x_{t} | \mu_{i_{e}}, \Sigma_{i_{e}})$.
In this case, $n_{i_{e}}$ is an index that takes values of up to $N_{i_{e}}$, which is the number of candidate points sampled for a specific $i_{e}$.
By sampling multiple points according to the Gaussian distribution, candidate waypoints that follow the rough shape of the place can be selected.
For example, the robot does not necessarily have to go to the center of a long corridor.

Therefore, as a concrete solution to Eq.~(\ref{eq:spconavi-HP:kai}), the partial path in the transition of candidate points from place $i_{e-1}$ to place $i_{e}$ is estimated as follows:
\begin{align}
    \hat{\mathbf{x}}_{i_{e-1},i_{e}}^{[ n_{i_{e-1}}, n_{i_{e}} ]}
    &= {\rm A}^{\star}(\hat{x}_{t'_{e-1}|i_{e-1}}^{[n_{i_{e-1}}]}, \hat{x}_{t'_{e}|i_{e}}^{[n_{i_{e}}]}, w_{e}),
    \label{eq:spconavi-HP:approx_A_star2}
\end{align}
where ${\rm A}^{\star}(s, g, w_{e})$ denotes the function of the A$^{\star}$ search algorithm, the initial position is $ s $, the goal position is $ g $, and the cost function is $w_{e}={\mathcal N}(x_{t} | \mu_{i_{e}}, \Sigma_{i_{e}}) p(x_{t} | m)$.
The estimated partial path length can be interpreted as the estimated value of $D_{e}$.

The selection of a series of partial metric path candidates corresponds to the selection of the entire path. 
Thus, we can replace the formulation of the maximization problem of Equation~(\ref{eq:spconavi-HP_detail3}) with that of Equation~(\ref{eq:spconavi-HP:jyoui2}).
Each partial metric path has corresponding indices $i_{e-1}$ and $i_{e}$.
Therefore, given a series of index pairs representing transitions between position distributions, the candidate paths to be considered can naturally be narrowed down to a series of corresponding partial paths.
The series of candidate indices that determines the series of candidate paths, in this case, is $\mathbf{n}_{1:E} = (n_{i_{0}}, n_{i_{1}}, \dots, n_{i_{E}})$.
This partial path sequence can be regarded as a sampling approximation of $x_{1:T}$.

By taking the maximum value instead of summing for $i_{1:E}$, path planning at the topological level is described as 
\begin{align}
    \mathbf{n}_{1:E}, i_{1:E}   
    &= \argmax_{\mathbf{n}_{1:E}, i_{1:E}}  \prod_{e=1}^{E} \frac{ {\rm Mult}(i_{e} \mid  \psi_{i_{e-1}}) }{ \sum_{c'=1}^{L} {\rm Mult}(i_{e} \mid \phi_{c'}) } \hat{\mathbf{w}}_{i_{e-1},i_{e}}^{[ n_{i_{e-1}}, n_{i_{e}} ]}  \nonumber \\ 
    & \qquad \sum_{{C}_{e}=1}^{L} {\rm Mult}(i_{e} \mid \phi_{C_{e}}) {\rm Mult}(S_{e} \mid W_{C_{e}}) {\rm Mult}(C_{e} \mid \pi),  
    \label{eq:spconavi-HP:jyoui2}
\end{align}
where $\hat{\mathbf{w}}_{i_{e-1},i_{e}}^{[ n_{i_{e-1}}, n_{i_{e}} ]}$ is the likelihood of the metric path $\hat{\mathbf{x}}_{i_{e-1},i_{e}}^{[ n_{i_{e-1}}, n_{i_{e}} ]}$ when transitioning from a candidate point of place $i_{e-1}$ to a candidate point of place $i_{e}$ at Step $e$.
In this case, it is equivalent to formulating the state variables in the distribution for the CaI as $\mathbf{x}_{1:E}$ and $i_{1:E}$.
Therefore, path planning at the topological level can be expressed as CaI at event step $e$.

\section{Experiment I: Planning tasks in simulator}
\label{sec:exp1}

We experimented with path planning using spatial concepts, including topological structures from human speech instructions.
In this experiment, as a first step, we demonstrated that the proposed method can improve the efficiency of path planning when the ideal spatial concept is in place.
The simulator environment was SIGVerse Version 3.0~\citep{inamura2021sigverse}. 
The virtual robot model in SIGVerse is the Toyota Human Support Robot (HSR). 
We used five three-bedroom home environments\footnote{three-dimensional (3D) home environment models are available at \url{https://github.com/a-taniguchi/SweetHome3D_rooms}.} with different layouts and room sizes. %

\subsection{Spatial concept-based topometric semantic map} 
\label{sec:exp1:learning_result}

\begin{figure}[!tb]
    \centering
    \includegraphics[width=0.60\linewidth, clip]{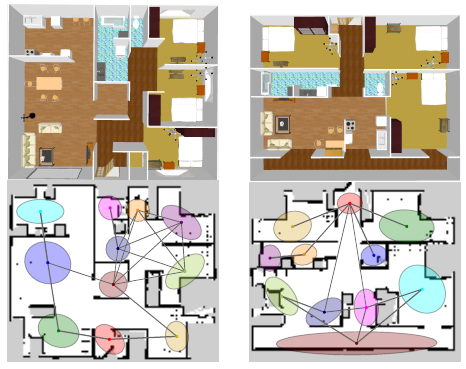} 
    \caption{Overhead view of the simulator environments (top); and the ideal spatial concepts expressed by SpCoTMHP on the environmental map (bottom), in Experiment I. 
    The colors of the position distributions were randomly set. 
    If $ ( \psi_{k_1,k_2} + \psi_{k_1,k_2} ) / 2 > 1/K $, the centers $ \mu_{k_1}, \mu_{k_2} $ of Gaussian distributions are connected by an edge.
    This means that the edges are drawn only if the average transition probability from {$k_{1}$} to {$k_{2}$} and {$k_{2}$} to {$k_{1}$} is higher than the uniform transition probability.}
    \label{fig:learning_sim}
\end{figure}

There were 11 spatial concepts and position distributions for each environment (See Fig.~\ref{fig:learning_sim} bottom and Appendix~\ref{apdx:exp1:env}).
Fifteen utterances were provided by the user for each place as training data.
The SLAM and speech recognition modules were inferred individually by splitting from the model; i.e., the self-location $x_{1:E}$ and word sequence $S_{1:E}$ were inputted into the model as observations.
An environment map was generated by the {\tt gmapping} package, which implements grid-based FastSLAM 2.0~\citep{gridbasedfastslam2007}, in the robot operating system (ROS). 
In this experiment, a word dictionary was prepared in advance for the vocabulary to be used, considering the focus was on evaluating path planning.
In addition, we assumed that the speech recognition result was obtained accurately.
Model parameters for the spatial concept were obtained via sampling from conditional distribution, i.e., Eq.~(\ref{eq:spco_gibbs}). 
We adopted the ideal learning results of spatial concepts and the latent variables $C_{t}$ and $i_{t}$ were accurately obtained.
Figure~\ref{fig:learning_sim} presents two examples of the overhead views of the home environments built on the simulator and the spatial concepts (i.e., the position distributions and their connections) in the environmental maps.

%

\subsection{Path planning from speech instructions}
\label{sec:exp1:path_planning}

Two types of path-planning tasks were performed in the experiment. 
This experiment was conducted in a variation where waypoints and goals were recombined at different places.
Waypoint and goal words in user instructions are extracted by any simple natural language process and entered into the model as $\{ S_{e} \}$.
\\
\textbf{Basic task:}
The robot obtained the word that identifies the target location as the instruction, e.g., ``{\it Go to the bedroom.}''
\\
\textbf{Advanced task:}
The robot obtained the words that identify the waypoint locations and target as the speech instruction, such as ``{\it Go to the bedroom via the corridor.}''
We inputted both the waypoint and target words as bag-of-words into SpCoNavi, as the task was not demonstrated in the previous work~\citep{ataniguchi2020spconavi}.

We compared the performances of the methods as follows: 
\begin{enumerate}[(A) ]
\item A$^\star$ algorithm (goal estimated by spatial concepts):
The goal position was obtained as $x^{\ast} \sim p(x \mid S^{\ast}, \Theta)$ in SpCoSLAM, using the speech recognition result $S^{\ast}$.
\item SpCoSLAM~\citep{ataniguchi_IROS2017} + SpCoNavi~\citep{ataniguchi2020spconavi} with Viterbi~\citep{viterbi1967error}
\item SpCoSLAM~\citep{ataniguchi_IROS2017} + SpCoNavi~\citep{ataniguchi2020spconavi} with A$^\star$ approx.
\item Hierarchical path planning (HPP) without CaI, similar to \citet{Niijima2020}: 
Goal node is estimated by $i_{e} \sim p(i_{e} \mid S^{\ast}, \mathbf{\Theta})$. The topological planning uses heuristic costs: (I) the cumulative cost and (II) the distance of partial paths in A$^\star$. 
\item SpCoTMHP (Topological: Dijkstra, Metric: A$^\star$) 
\end{enumerate}

The evaluation metrics for path planning represent the success weighted by path length (SPL)~\citep{Anderson2018} when the robot reaches the target location and calculation runtime seconds (time).
The N-SPL is the weighted success rate at which the robot reaches the closest target from the initial position when several places have the same name.
The W-SPL is the weighted success rate at which the robot passed the correct waypoints.
The WN-SPL is the weighted success rate at which the robot reaches the closest target and passes the correct waypoints.
The WN-SPL is an overall measure of path-planning efficiency in advanced tasks.

\textbf{Condition:}
The planning horizons were $E=10$ for the topological level and $D=100$ as the maximum limit value for the metric level in the SpCoTMHP.
The number of position candidates in the sample was $N_{i_{e}}=1$\footnote{This means a one-sample approximation to the candidate waypoints for the partial path. A related description can be found in Section~{\ref{sec:SpCoNavi-HP:approx}}.
A one-sample approximation will be sufficient if the Gaussian distributions representing the locations and their transitions are accurately obtained.}.
The proposed method subjected paths to moving average smoothing with a window size of 5.
The planning horizon of SpCoNavi was $T=200$.
The number of goal candidates of SpCoNavi (A$^\star$ approx.) was $J=10$.
The parameters $E$, $D$, and $T$ were large enough for the complexity of the environment.
$J$ is the same as in the original experimental setting~\citep{ataniguchi2020spconavi}.
The global cost map was obtained from the {\tt costmap\_2d} package in the ROS. 
The robot's initial position was set from arbitrary movable coordinates on the map. 
The user provided a word to indicate the target name.
The state of self-position $x_{t}$ is expressed discretely for each movable cell in the occupancy grid map $m$. 
The motion model assumes a simple deterministic model, i.e., {$x_t = x_{t-1} + u_{t}$}. 
In other words, motion errors are not assumed in path planning.
The control value $u_{t}$ assumes to move one cell on the map per time-step. 
Action $u_{t}$ is discretized into $\mathcal{A} = $ \{{\tt stay}, {\tt up}, {\tt down}, {\tt left}, {\tt right}\}.
This study was implemented using Python on one central processing unit (CPU), i.e., an Intel Core i7-6850K with 16 GB DDR4 2133-MHz synchronous dynamic random-access memory (SDRAM).

\begin{table*}[tb]
\begin{center}
\caption{Evaluation results of the path planning tasks in basic task (Experiment I).}
\begin{tabular}{lccccc} \hline
\textbf{Methods} & \textbf{Hierarchy} & \textbf{CaI} 
& \textbf{SPL}$\uparrow$ & \textbf{N-SPL}$\uparrow$ & \textbf{Time}$\downarrow$ 
\\ \hline
A$^{\star}$                             & - & - & 0.570                      & 0.463                      & $9.47 \times 10^{0}$
                                        \\ 
SpCoNavi  (Viterbi)                     & - & \checkmark & \underline{\textbf{0.976}} & \underline{\textbf{0.965}} & $2.68 \times 10^{3}$             
                                        \\ 
SpCoNavi  (A$^{\star}$ approx.)  & - & \checkmark & 0.404                      & 0.388                      & $5.42 \times 10^{1}$
                                        \\ 
HPP-I (path cost)           & \checkmark & - & 0.723                      & 0.605                      & \underline{$7.56 \times 10^{0}$} 
                                        \\ 
HPP-II (path distance)      & \checkmark & - & 0.714                      & 0.571                      & $7.96 \times 10^{0}$
                                        \\ 
SpCoTMHP                                 & \checkmark & \checkmark & \underline{0.861}          & \underline{0.812}          & \underline{$\mathbf{4.79 \times 10^{0}}$} 
                                        \\ 
\hline 
\end{tabular}
\label{tbl:hyouka_exp2_plan}
\end{center}
\end{table*}
\begin{table*}[tb]
\begin{center}
\caption{Evaluation results of the path planning tasks in advanced task (Experiment I).}
\begin{tabular}{lccccccc} \hline
\textbf{Methods} & \textbf{Hierarchy} & \textbf{CaI} 
& \textbf{SPL}$\uparrow$ & \textbf{W-SPL}$\uparrow$ & \textbf{N-SPL}$\uparrow$ & \textbf{WN-SPL}$\uparrow$  & \textbf{Time}$\downarrow$ \\ \hline
A$^{\star}$                             & - & - 
                                        & 0.312 & \underline{0.449}	& 0.233	& 0.034 & $9.44 \times 10^{0}$ \\ 
SpCoNavi  (A$^{\star}$ approx.)  & - & \checkmark 
                                        & 0.266 & 0.308	& 0.252	& 0.013 & $5.53 \times 10^{1}$ \\ 
HPP-I (path cost)           & \checkmark & - 
                                        & \underline{{0.917}} & 0.248	& \underline{0.773}	& \underline{0.191} & \underline{$7.53 \times 10^{0}$} \\ 
HPP-II (path distance)      & \checkmark & - 
                                        & {0.902} & 0.250	& 0.729	& 0.183 & $8.03 \times 10^{0}$ \\ 
SpCoTMHP                                 & \checkmark & \checkmark
                                        & \underline{\textbf{0.922}} & \underline{\textbf{0.906}}	& \underline{\textbf{0.794}}	& \underline{\textbf{0.781}} & \underline{$\mathbf{0.39 \times 10^{0}}$} \\ 
\hline 
\end{tabular}
\label{tbl:hyouka_exp2_plan_advanced}
\end{center}
\end{table*}
\begin{figure}[!tb]
  \centering
  \includegraphics[width=0.60\linewidth, clip]{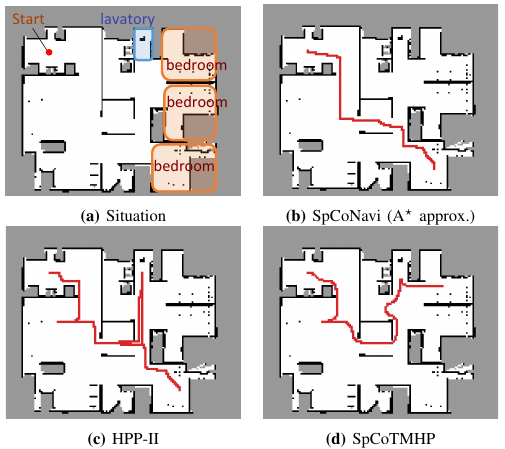} 
  \caption{Example of path planning in the advanced task.
  The instruction: ``{\it Go to the \textbf{bedroom} via the \textbf{lavatory}.}''
  (Experiment I).
  }
  \label{fig:exp1_plan}
\end{figure}
%

\textbf{Result:}
Tables~\ref{tbl:hyouka_exp2_plan} and~\ref{tbl:hyouka_exp2_plan_advanced} present the evaluation results for the basic and advanced planning tasks.
Figure~\ref{fig:exp1_plan} presents an example of the estimated path\footnote{A video of the robot simulation moving along the estimated path is available at {\url{https://youtu.be/w8vfEPtnWEg}}.}.
Overall, SpCoTMHP outperformed the comparison methods and significantly reduced the computation time.
The basic task demonstrated that the proposed method solves the problem of stopping the path before reaching the objective, which occurs in SpCoSNavi (A$^\star$ approx.).
The N-SPL of the baseline methods was lower than that of the proposed method because there were cases wherein the goal was selected as a bedroom far from the initial position (Fig.~\ref{fig:exp1_plan}~(b) and (c)).
This demonstrated the effectiveness of the proposed method based on probabilistic inference (i.e., CaI).

The advanced task confirmed that the proposed method could estimate the path via the waypoint (Fig.~\ref{fig:exp1_plan}~(d)).
Although SpCoTMHP had the disadvantage of estimating slightly redundant paths, the reduced computation time and improved planning performance render it a more practical approach than conventional methods.
Consequently, the proposed method achieved better path planning by considering all the initial, waypoint, and goal positions.

SpCoTMHP exhibited faster path planning compared to SpCoNavi (Viterbi), despite its inferior performance in basic task path planning. This improvement stems from the reduction in the number of inference states and computational complexity achieved through hierarchization and approximation. In both basic and advanced tasks, SpCoTMHP notably enhanced path planning performance compared to SpCoNavi (A$^\star$ approx.).
Consequently, the SpCoNavi problem outlined in Section~\ref{sec:preliminary_spco} was effectively addressed by SpCoTHMP.

\section{Experiment II: Real environment}
\label{sec:exp2}

We demonstrated that the formation of spatial concepts, including the topological relations of places, can be realized in a real-world environment.
Real-world datasets are more complex and involve more uncertainty than simulators.
Therefore, as detailed in Section~\ref{sec:exp2:learning_result}, we first confirmed that the proposed method could improve the learning performance when compared with the conventional method, SpCoSLAM.
Thereafter, as detailed in Section~\ref{sec:exp2:path_planning}, we determined the impact of the spatial concept parameters learned in Section~\ref{sec:exp2:learning_result} on the inference of path planning.
Additionally, we confirmed that the proposed method could plan a path based on the learned topometric semantic map.

\subsection{Spatial concept-based topometric semantic mapping}
\label{sec:exp2:learning_result}

\textbf{Condition:}
The experimental environment was identical to that in the open dataset {albert-b-laser-vision}{\footnote{Dataset is available in \url{https://dspace.mit.edu/handle/1721.1/62291}.}}, which was obtained from the robotics dataset repository (Radish)~\citep{Radish}. 
Details of the dataset are shown in Appendix~\ref{apdx:exp2:data}.
The utterance was 70 sentences in Japanese, such as ``\textit{The name of this place is student workroom},''  ``\textit{You can find the robot storage space here},'' and ``\textit{This is a white shelf}.'' 
The hyperparameters for learning were set as follows: $\alpha=0.5$, $\gamma=0.05$, $\beta=0.1$, $\chi=1.0$, $\omega=0.5$, $m_{0}=[ 0 , 0 ]^{\rm T}$, $\kappa_{0}=0.001$, $V_{0}={\rm diag}(2,2)$, and $\nu_{0}=3$. 
Parameters were set empirically within typical ranges with reference to SpCoSLAM~\citep{ataniguchi_IROS2017,ataniguchi2020spcoslam2}.
The other settings were the same as in Experiment~I.

\textbf{Evaluation metrics:}
Normalized mutual information (NMI)~\citep{kvalseth1987entropy} and adjusted Rand index (ARI)~\citep{hubert1985comparing}, which are the most widely used metrics in clustering tasks for unsupervised learning, were used as the evaluation metrics for learning the spatial concept.
Normalized mutual information is obtained by normalizing the mutual information between the clustering result and the correct label in the range 0.0--1.0.
Moreover, ARI is 1.0 when the clustering result matches the correct label and 0.0 when it is random.
Additionally, the time taken to learn was recorded as a reference value.

\begin{figure*}[tb]
\centering
\includegraphics[width=\linewidth, clip]{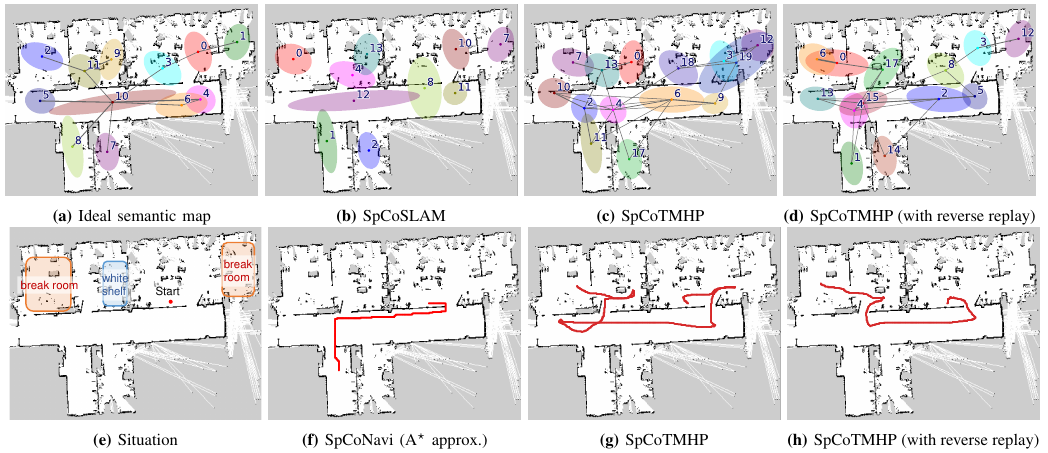} 
  \caption{Top (a--d): Result of the spatial concept learning. 
  Bottom (e--h): Result of path planning. 
  The speech instruction is ``{\it Go to the \textbf{break room} via the \textbf{white shelf}.}''
  The \textit{break room} was taught in two rooms: the upper right and the upper left. 
  The \textit{white shelf} is in the second room from the left on the upper side.
  (Experiment II).
  }
  \label{fig:plan_real}
\end{figure*}

\textbf{Result:}
Figures~\ref{fig:plan_real} (a--d) present an example of learning the spatial concept.
For example, Fig.~\ref{fig:plan_real} (c) caused overlapping distributions in the upper right and skipped connections to neighboring distributions, whereas (d) mitigated these problems.
Table~\ref{tbl:hyouka_real_learning} presents the results of evaluating the average values of ten trials of spatial concept learning.
SpCoTMHP achieved a higher learning performance (i.e., NMI and ARI values) than SpCoSLAM. 
This indicates that the categorization of spatial concepts and position distributions becomes more accurate on considering the connectivity of the place.
In addition, the proposed method with reverse replay demonstrated the highest performance.
Consequently, using both place transitions during learning (and vice versa) is useful for learning spatial concepts.
Moreover, Table~\ref{tbl:hyouka_real_learning} shows the computation time of the learning algorithm.
No significant difference was observed in the computation time.

\begin{table}[tb]
    \begin{center}
    \caption{Performance of learning for spatial concepts and position distributions and the computation time of the learning algorithm 
    (Experiment II).}
    \begin{tabular}{lccccc} \hline
     & \multicolumn{2}{c}{\textbf{NMI}$\uparrow$} & \multicolumn{2}{c}{\textbf{ARI}$\uparrow$} & {\textbf{Time}$\downarrow$} \\  
    \textbf{Methods} & $C_{e}$ & $i_{e}$ & $C_{e}$ & $i_{e}$ & {(sec.)}\\ \hline
    SpCoSLAM       & 0.767 & 0.803 & 0.539 & 0.578 & {$1.28 \times 10^{2}$} \\ 
    SpCoTMHP       & \underline{0.779}& \underline{0.858} & \underline{0.540} & \underline{0.656} & {$1.33 \times 10^{2}$} \\ 
    SpCoTMHP (with reverse replay) & \underline{\textbf{0.786}} & \underline{\textbf{0.862}} & \underline{\textbf{0.562}} & \underline{\textbf{0.658}} & {$1.29 \times 10^{2}$} \\ 
    \hline 
    \end{tabular}
    \label{tbl:hyouka_real_learning}
    \end{center}
\end{table}
%

\subsection{Path planning from speech instructions}
\label{sec:exp2:path_planning}

The speech instruction was ``{\it Go to the {break room} via the {white shelf}.}''
The other settings were the same as in Experiment~I.
%
Figures~\ref{fig:plan_real} (e--h) present the results for path planning using the spatial concept. 
Although SpCoSLAM could not reach the waypoint and goal \addspan{in (f)}, SpCoTMHP could estimate the path to reach the goal via the waypoint \addspan{in (g-h)}.
The learning with reverse replay in (\addspan{d}) shortened the additional route that would result from the bias of the transition between places during learning in (\addspan{c}). 
\addspan{
The failure observed in (f) with SpCoNavi using waypoints is primarily due to the inputting of the names of given locations, whether they are waypoints or goals, in a bag-of-words format. 
}
The results revealed that the proposed method can accurately perform hierarchical path planning, although the learning results are incomplete, as shown in Table~\ref{tbl:hyouka_real_learning}.
{As a reference value, inference time for path planning was $1.02 \times 10^{3}$ sec. for SpCoNavi, $3.97 \times 10^{-2}$ sec. for SpCoTMHP, and $2.39 \times 10^{-2}$ sec. for SpCoTMHP (with reverse replay).
In Experiment I (Section~\ref{sec:exp1}), the results demonstrate the computational efficiency of the proposed hierarchical path planning.
}

\section{Conclusions} 
\label{sec:conclusion}

We achieved topometric semantic mapping based on multimodal observations and hierarchical path planning through waypoint-guided instructions.
Experimental results demonstrated improved performance in spatial concept learning and path planning, both in simulated and real-world environments.
Additionally, the approximate inference achieved a high computational efficiency, countering the model's complexity.

However, our study has a few limitations, as follows:
\begin{enumerate}[1. ]
\item \addspan{\textbf{Scalability:}}
The experiment assumed one waypoint; however, the proposed method can theoretically handle multiple waypoints.
Although the computation complexity increases with the topological planning horizon $E$, scalability will be sufficiently ensured when users only require a few waypoints.
In practical scenarios, one or two waypoints are highly probable in daily life.
%
\item \addspan{\textbf{Instruction Variability:}}
A typical instruction representation was used in the experiment.
As a preprocessing step, using LLMs can handle instruction variability~\citep{Shah2022LM-Nav}.
\addspan{
\item \textbf{Redundant Waypoints:}
Our approach may require passing through redundant waypoints, even if visiting the waypoint itself is unnecessary. 
For instance, in Fig.~\ref{fig:plan_real}, if it were possible to directly specify ``the break room next to the white shelf,'' there would be no need to go through the white shelf as a waypoint. 
In such cases, extending the system to an open-vocabulary LLM-based semantic map could provide a solution.
}
\addspan{
\item \textbf{Path Restrictions:}
The paths generated by the proposed model are restricted based on the transition probabilities between locations encountered during training. 
In contrast, the model by \citet{Banino2018} can navigate through paths not traversed during training. 
Exploring the integration of such vector-based navigation techniques with our spatial concept-based approach could potentially enable shortcut navigation and enhance the model's flexibility and robustness.
}
\end{enumerate}

Future research will include utilizing common sense reasoning~\citep{Hasegawa2023} such as foundation models and transfer of knowledge~\citep{Katsumata2020SpCoMapGAN} with respect to spatial adjacencies across multiple environments. 
In this study, we trained the model using the procedure described in Sec.~\ref{sec:SpCoSLAM-TM:learning}.
Simultaneous and online learning for the entire model can be realized with particle filters~\citep{ataniguchi_IROS2017}.
The proposed method was found to be computationally efficient, thus potentially rendering it applicable to online path planning, e.g., model predictive control~\citep{Stahl2011,Li2019c}.
Additionally, the proposed model has the potential to enable visual navigation and the generation of linguistic path explanations through cross-modal inference by the robot.


\section*{Conflict of Interest Statement}

The authors declare that the research was conducted without any commercial or financial relationships that may be construed as a potential conflict of interest.

\section*{Author Contributions}
AT, SI, and TT conceived, designed the research, and wrote the paper.
AT experimented and analyzed the data.


\section*{Funding}
This work was partially supported by JST CREST under Grant number JPMJCR15E3, including the AIP Challenge Program, 
JST Moonshot Research \& Development Program under Grant number JPMJMS2011,
and JSPS KAKENHI under Grant numbers JP20K19900, JP21H04904, and JP23K16975.

\section*{Acknowledgments}
The authors thank Cyrill Stachniss for providing the {\tt albert-b-laser-vision} dataset. 
The authors also thank Kazuya Asada and Keishiro Taguchi for providing virtual home environments and training datasets for spatial concepts in the SIGVerse simulator.




\bibliographystyle{Frontiers-Harvard}
\bibliography{SpCoTMHP} 


\newpage

\appendix
\section*{Appendix}
\section{Formulation of the generative process of SpCoSLAM and SpCoNavi}
\label{apdx:SpCoSLAM:overview:formulation}
The details of the formulation of the generative process represented by the graphical model of SpCoSLAM and SpCoNavi can be described as follows:
\begin{eqnarray}
\pi       &\sim& {\rm DP}(\alpha ) \label{apdx:eq:seisei1} \\
\phi_{l}  &\sim& {\rm DP}(\gamma ) \label{apdx:eq:seisei5} \\
\theta_{l}&\sim& {\rm Dir}(\chi) \label{apdx:seisei3} \\
W_{l}     &\sim& {\rm Dir}(\beta) \label{apdx:eq:seisei3} \\
LM        &\sim&  p(LM \mid \lambda) \label{apdx:seiseilm} \\
\Sigma_{k}&\sim& {\mathcal IW}( V_{0}, \nu _{0} )  \label{apdx:eq:seisei7} \\
\mu_{k}   &\sim& {\mathcal N}( m_{0}, \Sigma_{k} / \kappa _{0} ) \label{apdx:eq:seisei8} \\
x_{t}     &\sim& p(x_{t} \mid x_{t-1},u_{t}) \label{apdx:eq:seisei9} \\
z_{t}     &\sim& p(z_{t} \mid x_{t},m) \label{apdx:eq:seisei10} \\
C_{t}     &\sim& {\rm Mult}(\pi) \label{apdx:eq:seisei2} \\
i_{t}     &\sim& p(i_{t} \mid x_{t}, \mbox{\boldmath{$\mu $}}, \mbox{\boldmath{$\Sigma$}}, \mbox{\boldmath{$\phi$}}, C_{t}) \label{apdx:eq:seisei6} \\
f_{t}     &\sim& {\rm Mult}(\theta_{C_{t}}) \label{apdx:seisei2} \\
S_{t}     &\sim& p(S_{t} \mid {\bf{W}}, C_{t},LM) \label{apdx:seisei4b} \\
y_{t}     &\sim& p(y_{t} \mid S_{t},AM) \label{apdx:seiseiYt} 
\label{apdx:eq:seisei}
\end{eqnarray}
where ${\rm DP()}$ represents the Dirichlet process, ${\rm Dir()}$ is the Dirichlet distribution, ${\mathcal IW()}$ is the inverse–Wishart distribution, ${\mathcal N}()$ is the multivariate normal distribution, and ${\rm Mult()}$ is the multinomial distribution.
See \citep{murphy2012machine} for the specific formulas of the above probability distributions.

The probability distribution of Equation~(\ref{apdx:eq:seisei9}) represents a motion model, i.e., a state transition model, in SLAM.
The probability distribution of Equation~(\ref{apdx:eq:seisei10}) represents a measurement model in SLAM.

The probability distribution of Equation~(\ref{apdx:eq:seisei6}) can be defined as 
\begin{eqnarray}
p(i_{t} \mid x_{t},\mbox{\boldmath{$\mu $}},\mbox{\boldmath{$\Sigma$}} ,\mbox{\boldmath{$\phi$}}, C_{t})
&=& \cfrac{{\mathcal N}(x_{t} \mid \mu _{i_{t}}, \Sigma_{i_{t}}){\rm Mult}(i_{t} \mid \phi_{C_{t}})}{\sum_{i_{t}=j} {\mathcal N}(x_{t} \mid \mu _{j}, \Sigma_{j}){\rm Mult}(j \mid \phi_{C_{t}})}~.
\label{eq:it}
\end{eqnarray}

The probability distribution of Equation~(\ref{apdx:seisei4b}) is approximated by unigram rescaling~\citep{gildea1999topic}, as 
\begin{eqnarray}
p(S_{t} \mid {\bf{W}}, C_{t},LM)&{\approx }& p(S_{t} \mid LM) \prod_{B_{t}} \frac{{\rm Mult}(S_{t,b} \mid W_{C_{t}})}{\sum_{c'}{\rm Mult}(S_{t,b} \mid W_{c'})},
\label{apdx:st_UR}
\end{eqnarray}
where $B_{t}$ denotes the number of words in the sentence and $S_{t,b}$ is $b$-th word in the sentence at the time-step of $t$.

\section{Formulation and procedure for each step of the online learning algorithm}
\label{apdx:SpCoSLAM:learning:procedure}
The online learning algorithm introduces sequential equation updates to estimate the parameters of the spatial concepts into the formulation of a Rao–Blackwellized particle filter~\citep{doucet2000rao} in the FastSLAM~2.0~\citep{montemerlo2003fastslam} and its grid-based SLAM~\citep{gridbasedfastslam2007}.
The particle filter is advantageous, in that parallel processing can be readily applied because the calculations concerning the particles can be calculated independently.
{Theoretically, other particle-filter-based SLAMs, besides FastSLAM 2.0, can also be used.}

In the formulation of SpCoSLAM, the joint posterior distribution can be factorized to the probability distributions of a language model $LM$, a map $m$, the set of model parameters of spatial concepts $\Theta = \{ {\mathbf W}, \mbox{\boldmath $\mu $}, \mbox{\boldmath $\Sigma$}, {\mathbf \theta}, {\mathbf \phi}, \pi \}$, the joint distribution of the self-positions $x_{0:t}$, and the set of latent variables $\mathbf{C}_{1:t} = \{i_{1:t},C_{1:t},S_{1:t} \}$.
The joint posterior distribution can be described as follows:
\begin{eqnarray}
&&p(x_{0:t},\mathbf{C}_{1:t}, LM, \Theta, m 
\mid u_{1:t}, z_{1:t}, y_{1:t}, f_{1:t}, AM ,\mathbf{h}) \nonumber \\
&&=p(LM \mid S_{1:t}, \lambda)
p(\Theta \mid x_{0:t}, \mathbf{C}_{1:t}, f_{1:t}, \mathbf{h})p(m \mid x_{0:t}, z_{1:t}) \nonumber \\
&&\hspace{1.0em}\cdot~\underbrace{p(x_{0:t},\mathbf{C}_{1:t} \mid u_{1:t}, z_{1:t}, y_{1:t}, f_{1:t}, AM ,\mathbf{h})}_\text{Particle~filter}
\label{apdx:eq:spcoslam}
\end{eqnarray}
where the set of hyperparameters is denoted by $\mathbf{h}= \{ \alpha,\beta,\gamma,\chi,\lambda, m_{0},\kappa_{0}, V_{0},\nu_{0} \}$.

The variables of the joint posterior distribution can be learned by Gibbs sampling, which is a Markov chain Monte-Carlo-based batch learning algorithm, in a manner similar to the nonparametric Bayesian spatial concept acquisition method (SpCoA)~\citep{taniguchi_spcoa}.
{In addition, learning can be realized in a spatial concept formation model after the map is generated via any other SLAM.}

The learning procedure of SpCoSLAM for each step is described as follows:
\begin{enumerate}[(a) ]
\item The robot obtains weighted finite-state transducer (WFST) speech recognition results $\mathcal{L}_{1:t}$ from the user speech signals $y_{1:t}$ using a language model $LM$. 
The WFST is a word graph, i.e., a lattice format, which alternatively represents the $N$-best speech recognition results.
Initially, a phoneme dictionary is provided as the language model $LM$ without a prior word list.
\item The WFST speech recognition results $\mathcal{L}_{1:t}$ are segmented to the word sequences $S_{1:t}$ using an unsupervised word segmentation approach referred to as latticelm~\citep{neubig2012bayesian}. 
\item The latent variable $x_{t}$ and importance weight $\omega_{z}$ regarding self-localization are obtained by the grid-based FastSLAM 2.0 {from control data $u_{t}$, depth data $z_{t}$ and particles that represent the self-positions $x_{t-1}$ at the previous time step}.
\item The latent variables $ i_{t}, C_{t} $ of spatial concepts are sampled by the proposal distribution on the particle filter.
\item The importance weights $\omega_{s}$, $\omega_{f}$ are obtained as the marginal likelihoods of observations $ S_{t}, f_{t}$.
\item The environmental map $m$ is updated by self-positions $x_{0:t}$ and depth data $z_{1:t}$.
\item The set of model parameters $\Theta$ of the spatial concepts are estimated from the observation $f_{1:t}$ and sampled variables $ x_{0:t}, \mathbf{C}_{1:t}$. 
\item The language model $LM$ is updated by adding words $S_{1:t}^{\ast}$ in a particle of maximum weight to the initial dictionary.
\item The particles are re-sampled according to their weights $\omega_{t}= \omega_{z} \cdot \omega_{s} \cdot \omega_{f}$. 
\end{enumerate}
Steps (b) -- (g) are performed for each particle.
See the original paper~\citep{ataniguchi_IROS2017} for details.

\section{Control as probabilistic inference}
\label{apdx:SpCoNavi:CaI}

The theoretical gap between the control problems, including reinforcement learning (RL) and the probabilistic inference in the generative model, was bridged by CaI~\citep {levine2018reinforcement}.

\begin{figure}
    \begin{center}
        \includegraphics[width=0.7\textwidth]{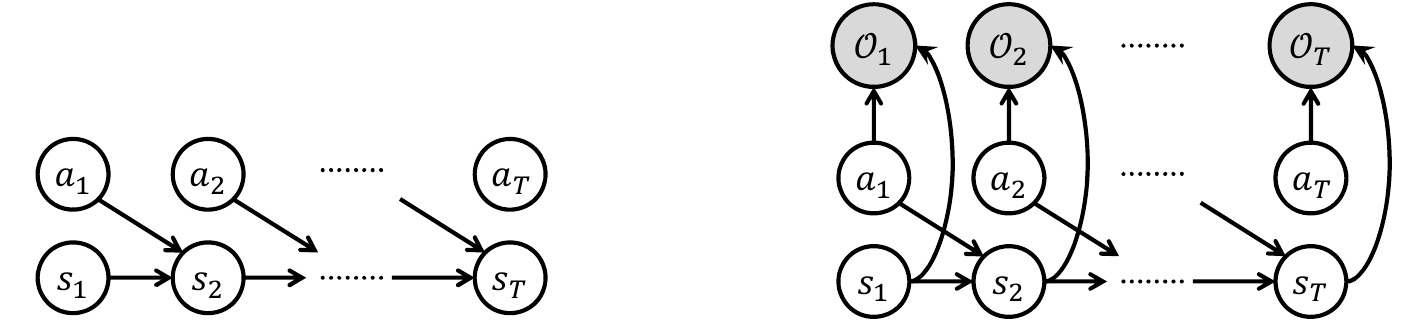}
        \caption{
            Left: Graphical model of Markov decision process (MDP) with states and actions. 
            Right: Graphical model for CaI with the optimality variables. 
            This additional variable is a binary random variable, where ${\mathcal{O}}_{t}=1$ denotes that the time-step $t$ is optimal.
        }
        \label{fig:CaI}
    \end{center}
\end{figure}

In general decision-making problems, including RL, a policy that maximizes the expected cumulative reward is estimated as follows:
\begin{eqnarray}
\vartheta^{\star} &=& \argmax_{\vartheta} \sum_{t=1}^{T}  \mathbb{E}_{(s_{t}, a_{t}) \sim p(s_{t},a_{t} \mid \vartheta)} \left[ r(s_{t}, a_{t}) \right],
\label{eq:expected_reward}
\end{eqnarray}
where $r(s_{t}, a_{t})$ is a reward function, $s_{t}$ is a state variable, $a_{t}$ is an action variable, $\vartheta$ is a parameter for the policy function, and $\vartheta^{\ast}$ is an optimal policy parameter.
It should be noted that $s_{t}$ and $a_{t}$ correspond to $x_{t}$ and $u_{t}$ in our method, respectively.

With respect to CaI, the planning problem can be formulated as an inference from the probabilistic graphical model. 
Figure~\ref{fig:CaI} presents the graphical models of the Markov decision process (MDP) with an optimality variable ${\mathcal{O}}_{t}$. 
In the graphical model for CaI, the distribution denoting the generative process on the binary random variable ${\mathcal{O}}_{t}$ is represented as 
\begin{eqnarray}
p({\mathcal{O}}_{t}=1 \mid s_{t}, a_{t}) &=& \exp(r(s_{t}, a_{t})).
\label{eq:bainary}
\end{eqnarray}

The maximum a posteriori inference in the posterior distribution $p(\tau \mid o_{1:T})$ corresponds to a type of planning problem. 
Here, trajectory is $\tau=\{ s_{1:T}, a_{1:T} \}$ and the set of optimality variables is $o_{1:T} = \{ {\mathcal{O}}_{t}=1 \}^{T}_{t=1}$.
The posterior distribution over actions when we condition ${\mathcal{O}}_{t}=1$ for all $t \in \{ 1, \dots, T\}$ is shown as 
\begin{eqnarray}
p(\tau \mid o_{1:T})
&\propto& p(s_{1})\prod_{t=1}^{T} p(s_{t+1} \mid s_{t}, a_{t}) p({\mathcal{O}}_{t}=1 \mid s_{t}, a_{t}) \nonumber \\
&=& \underbrace{ \left[ p(s_{1})\prod_{t=1}^{T} p(s_{t+1} \mid s_{t}, a_{t}) \right] }_\text{State-transition with action} \exp \underbrace{ \left( \sum_{t=1}^{T} r(s_{t}, a_{t}) 
\right) }_\text{Cumulative reward}. 
\label{eq:posterior}
\end{eqnarray}
This suggests that the optimalities $o_{1:T}$ are given as observations in a HMM-style model.
Therefore, the trajectory probability is given by the product between its probability to occur according to the dynamics and the exponential of the cumulative reward along that trajectory.

In this case, the policy function can be expressed as $\pi_{\vartheta}(s_{t}, a_{t}) = p(a_{t} \mid s_{t}, \vartheta)$.
The optimal policy function can be expressed as $p(a_{t} \mid s_{t}, \vartheta^{\star}) \approx p(a_{t} \mid s_{t}, {o}_{t:T})$; then, the right side is not related to the parameter $\vartheta$.

\citet{levine2018reinforcement} has described that the CaI allows for the application of various techniques of probabilistic inference, e.g., the forward–backward algorithm and variational inference, for control and planning problems.

\section{Simulator environments and their spatial concepts in Experiment I}
\label{apdx:exp1:env}

\begin{figure}[tb]
	\centering
       	\includegraphics[width=0.19\linewidth]{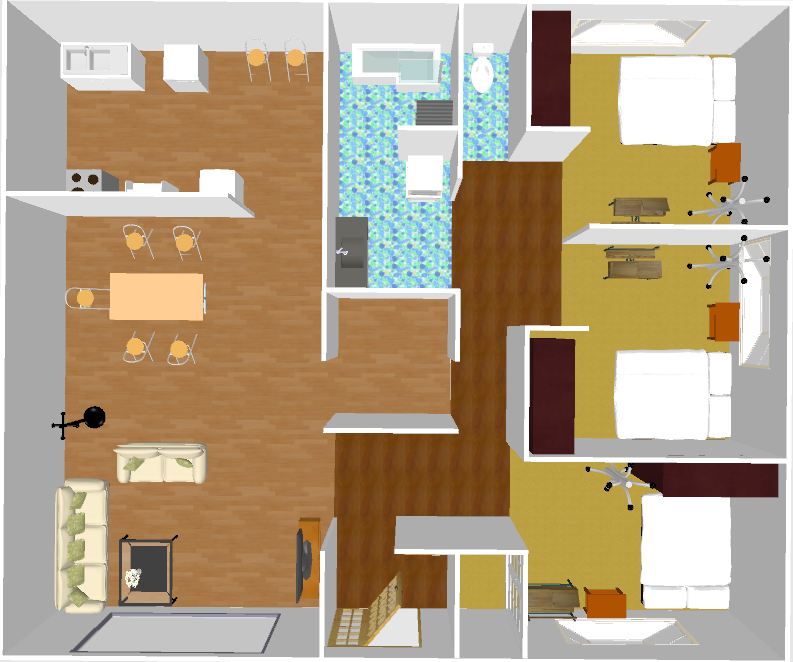}
       	\includegraphics[width=0.21\linewidth]{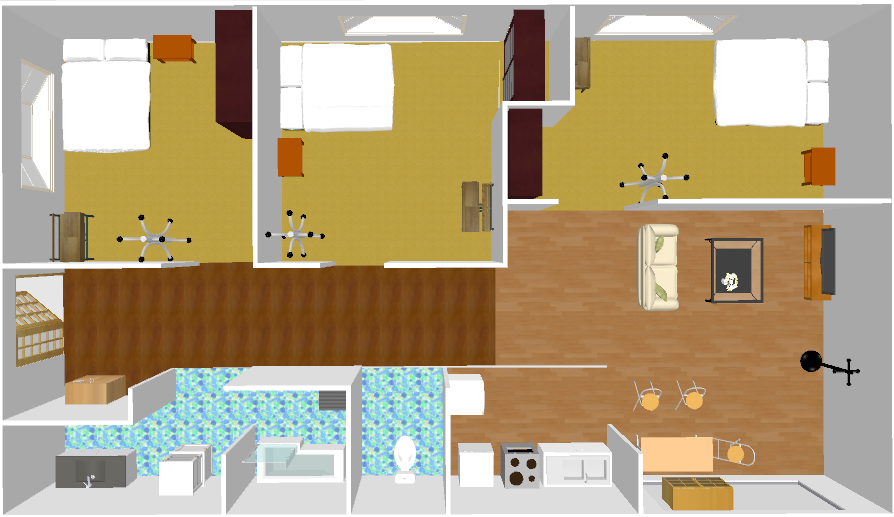}
       	\includegraphics[width=0.17\linewidth]{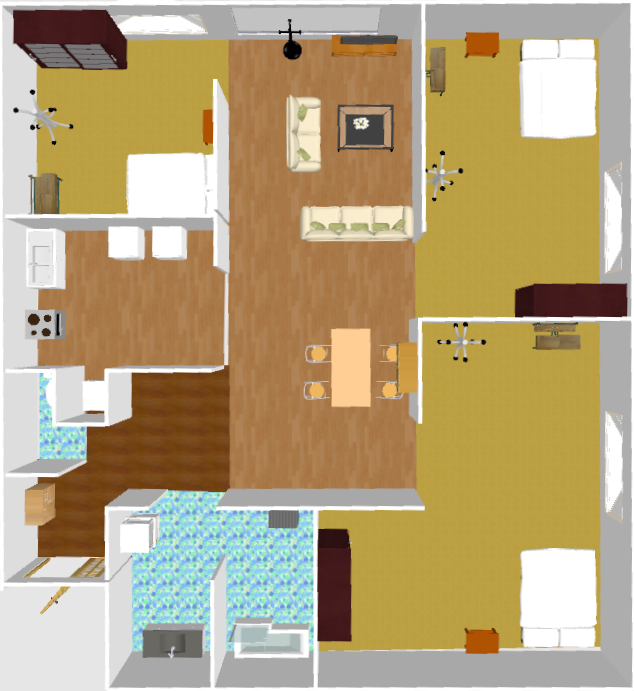}
       	\includegraphics[width=0.18\linewidth]{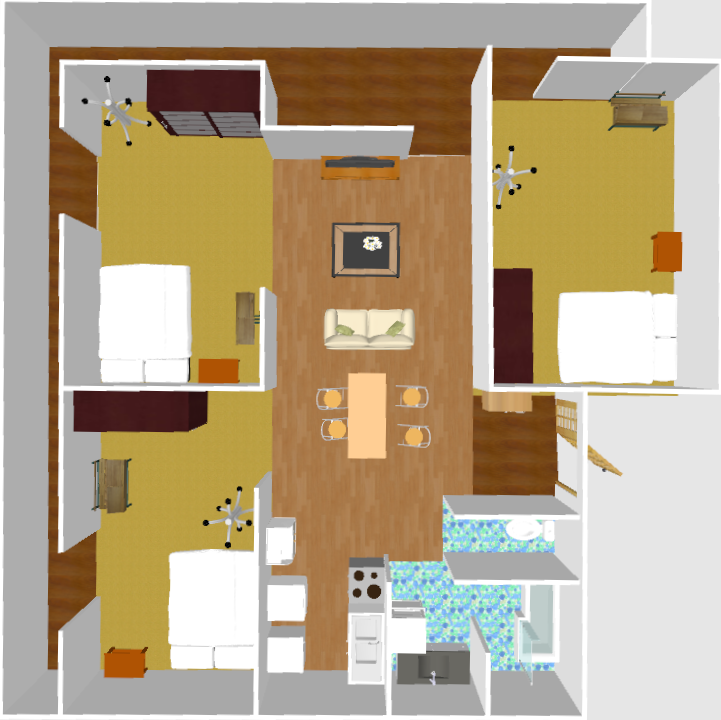}
       	\includegraphics[width=0.20\linewidth]{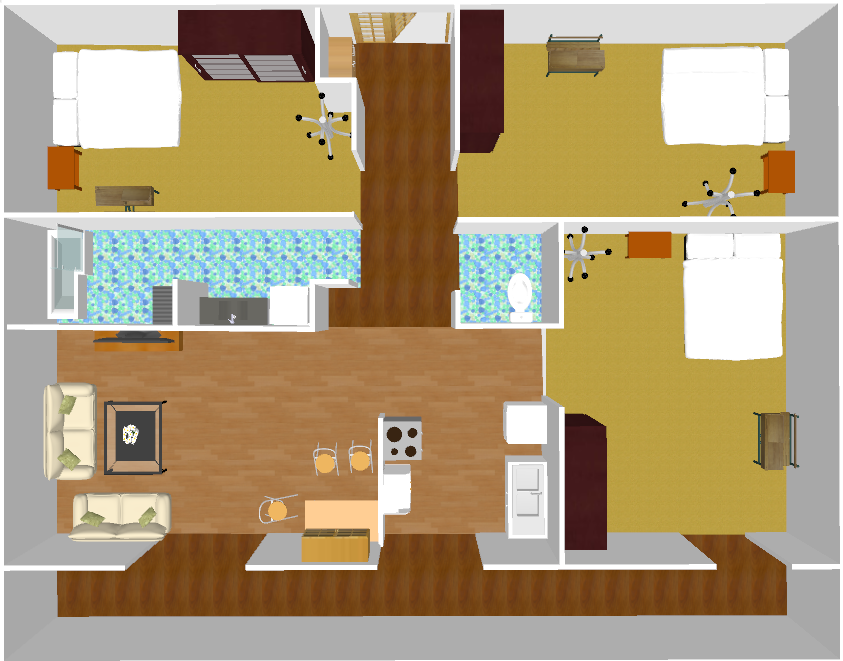}
	\caption{{Home environments created with SIGVerse in Experiment I}}
	\label{fig:exp_1_environment_5}
\end{figure}

\begin{figure}[tb]
	\centering
       	\includegraphics[width=0.19\linewidth]{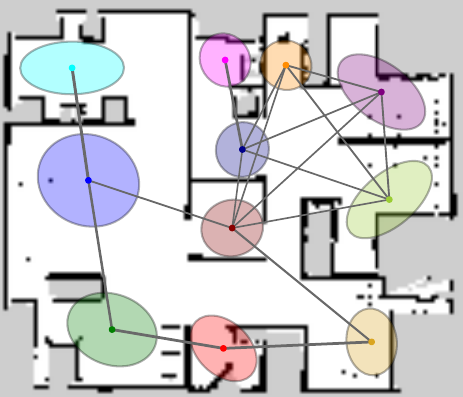}
       	\includegraphics[width=0.21\linewidth]{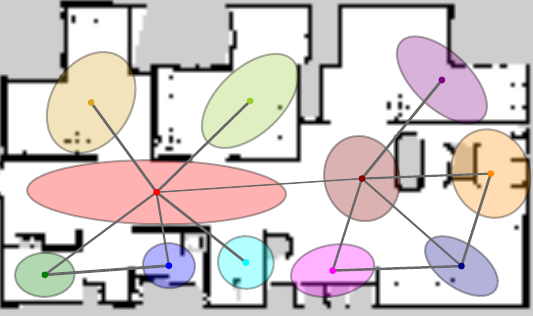}
       	\includegraphics[width=0.17\linewidth]{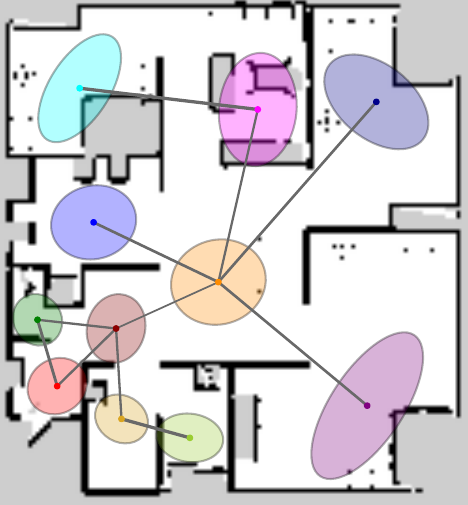}
       	\includegraphics[width=0.18\linewidth]{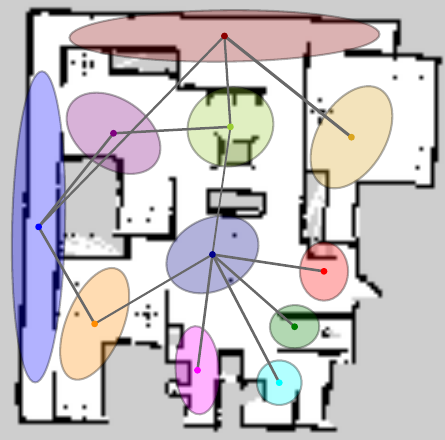}
       	\includegraphics[width=0.20\linewidth]{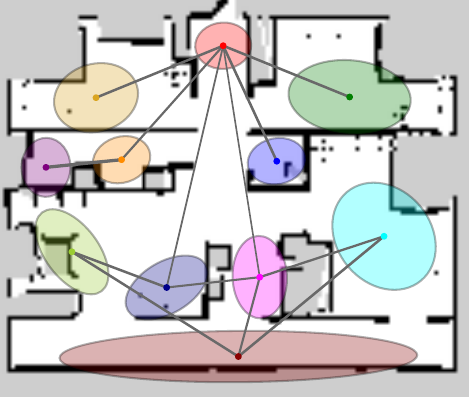}
	\caption{{Spatial concepts on home environments in Experiment I}}
	\label{fig:exp_1_spatial_concept}
\end{figure}

This section introduces the simulated home environments generated in SIGVerse and outlines the spatial concepts explored in Experiment I.
Figure~\ref{fig:exp_1_environment_5} shows the home environments.
Figure~\ref{fig:exp_1_spatial_concept} shows
It includes eleven spatial concepts and position distributions for each environment.
The terminology associated with each place varies depending on the environment. The list of words for spatial concepts is provided below:
``Entrance'', ``Living room'', ``Dining room'', ``Kitchen'', ``Bathroom'', ``Dressing room'', ``Lavatory'', ``Bedroom'', ``Storage space'', ``Corridor'', ``Open space'', ``East side'', and ``South side''

%
%

%

\section{Dataset of albert-b-laser-vision in Experiment II}
\label{apdx:exp2:data}

\begin{figure*}[tb]
\centering
\includegraphics[width=0.4\linewidth, clip]{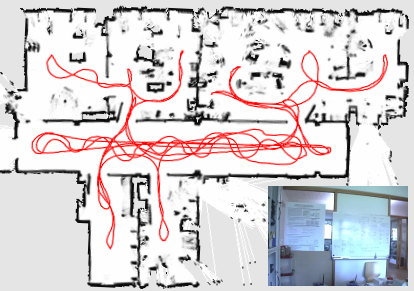} 
  \caption{{
  The robot trajectory within the dataset for learning is depicted by a red curved line, while an example image captured by the robot is shown in Experiment II.
  This image is included in the dataset~\citep{Radish}.
  }
  }
  \label{fig:albartB_data}
\end{figure*}

{
The experimental environment matched that of the open dataset {albert-b-laser-vision}, sourced from the robotics dataset repository (Radish)~\citep{Radish}. 
This dataset was captured using the b21r robot Albert within Building 79, University of Freiburg. 
The dataset comprises a log file containing odometry, laser range data, and image data. Detailed information about the robot's properties is provided within the dataset.}
{
Figure~\ref{fig:albartB_data} shows the robot trajectory image of the dataset.
}

\end{document}